%% file: main.tex
\documentclass[lettersize,journal]{IEEEtran}
\usepackage{amsmath,amsfonts}
\usepackage{algorithmic}
\usepackage{algorithm}
\usepackage{array}
\usepackage[caption=false,font=normalsize,labelfont=sf,textfont=sf]{subfig}
\usepackage{textcomp}
\usepackage{stfloats}
\usepackage{url}
\usepackage{verbatim}
\usepackage{graphicx}
\usepackage{cite}
\usepackage{enumitem}
\usepackage{xcolor}
\usepackage{booktabs}
\usepackage{cellspace}
\usepackage{float}
\usepackage{multirow}
\usepackage{lscape}
\usepackage{booktabs}
\usepackage{makecell}
\usepackage{adjustbox}

\newcommand{\aka}{\textit{a.k.a. }}
\newcommand{\ie}{\textit{i.e.}}

\newcommand{\etal}{\textit{et al. }}

\newcolumntype{M}[1]{>{\centering\arraybackslash}m{#1}}
\newcolumntype{P}[1]{>{\centering\arraybackslash}p{#1}}

\begin{document}

\title{Road Surface Defect Detection -- From Image-based to Non-image-based: A Survey}

\author{Jongmin Yu$^{1}$, Jiaqi Jiang$^{2}$, Sebastiano Fichera$^{3,4}$, Paolo Paoletti$^{3,4}$, Lisa Layzell$^{4}$,\\ Devansh Mehta$^{4}$, and Shan Luo$^{2\dagger}$,~\IEEEmembership{Members,~IEEE}
\thanks{$^{1}$J. Yu is with the Department of Applied Mathematics and Theoretical Physics, University of Cambridge, Wilberforce Rd, Cambridge CB3 0WA, United Kingdom; Emails: jy522@cam.ac.uk (was affiliated with Department of Engineering, King's College London, Strand, London, WC2R 2LS, United Kingdom)}
\thanks{$^{1}$J. Jiang and S. Luo are with the Department of Engineering, King's College London, Strand, London WC2R 2LS, United Kingdom; Emails: \{jiaqi.1.jiang, shan.luo\}@kcl.ac.uk}
\thanks{$^{2}$S. Fichera and P. Paoletti are with the Department of Mechanical, Materials and Aerospace Engineering, School of Engineering, University of Liverpool, Liverpool, L69 3GH, United Kingdom; Emails: \{Sebastiano.Fichera, paoletti\}@liverpool.ac.uk}
\thanks{$^{3}$S. Fichera, P. Paoletti, L. Layzell and D. Mehta are with Robotiz3d Ltd, Daresbury, WA4 4FS, United Kingdom; Emails: \{sebastiano.fichera, paolo.paoletti, lisa.layzell, devansh.mehta\}@robotiz3d.com}
\thanks{This work was supported by the Innovate UK SMART grant ``ARRES PREVENT: The World-First Autonomous Road Repair Vehicle'' (10006122).}
\thanks{$\dagger$ represents the corresponding author.}
\thanks{Manuscript received April 19, 2023; revised xx xx, 2023.}}

\markboth{Journal of \LaTeX\ Class Files,~Vol.~XX, No.~X, XX~202X}%
{Shell \MakeLowercase{\textit{et al.}}: A Sample Article Using IEEEtran.cls for IEEE Journals}


\maketitle

\begin{abstract}
Ensuring traffic safety is crucial, which necessitates the detection and prevention of road surface defects. As a result, there has been a growing interest in the literature on the subject, leading to the development of various road surface defect detection methods. The methods for detecting road defects can be categorised in various ways depending on the input data types or training methodologies. The predominant approach involves image-based methods, which analyse pixel intensities and surface textures to identify defects. Despite popularity, image-based methods share the distinct limitation of vulnerability to weather and lighting changes. To address this issue, researchers have explored the use of additional sensors, such as laser scanners or LiDARs, providing explicit depth information to enable the detection of defects in terms of scale and volume. However, the exploration of data beyond images has not been sufficiently investigated. In this survey paper, we provide a comprehensive review of road surface defect detection studies, categorising them based on input data types and methodologies used. Additionally, we review recently proposed non-image-based methods and discuss several challenges and open problems associated with these techniques.
\end{abstract}


\input{01_Introduction}
\input{02_Road_defects_categories}

\input{025_Dataset}
\input{03_Method}
\input{05_Conclusion}

\bibliographystyle{ieeetr}

\bibliography{cas-refs}

\small
\begin{IEEEbiography}[{\resizebox{1in}{!}{\includegraphics[width=1in,height=1.25in,clip]{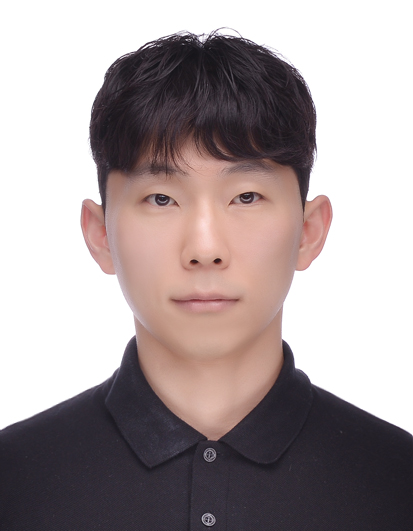}}}]{Jongmin Yu} is a postdoctoral research associate at the University of Cambridge. He was a research associate at the King's College London and Institute of IT Convergence at the Korea Advanced Institute of Science and Technology (KAIST). He received PhDs from the School of Electrical Engineering and Computer Science at Gwangju Institute of Science and Technology (GIST), Gwangju, Korea, Republic of, and the School of Electrical Engineering, Computing and Mathematical Sciences at Curtin University, Perth, Western Australia, Australia. His research interests include artificial intelligence, machine learning, pattern recognition, and mathematical understanding of these.
\end{IEEEbiography}

\small
\begin{IEEEbiography}[{\resizebox{1in}{!}{\includegraphics[width=1in,height=1.25in,clip]{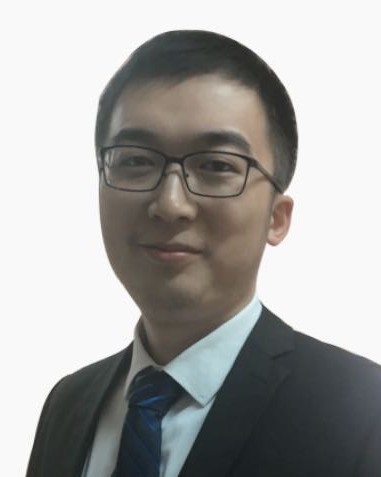}}}]{Jiaqi Jiang} received the B.S. and M.S. degrees from Beijing Institute of Technology, in 2016 and in 2019, respectively. He is currently a Ph.D. candidate in the Department of Engineering, King's College London. He was a Ph.D. candidate in the Department of Computer Science, the University of Liverpool. His research interests include robot grasping and sensory synergy of vision and touch.
\end{IEEEbiography}

\begin{IEEEbiography}[{\resizebox{1in}{!}{\includegraphics[width=1in,height=1.25in,clip]{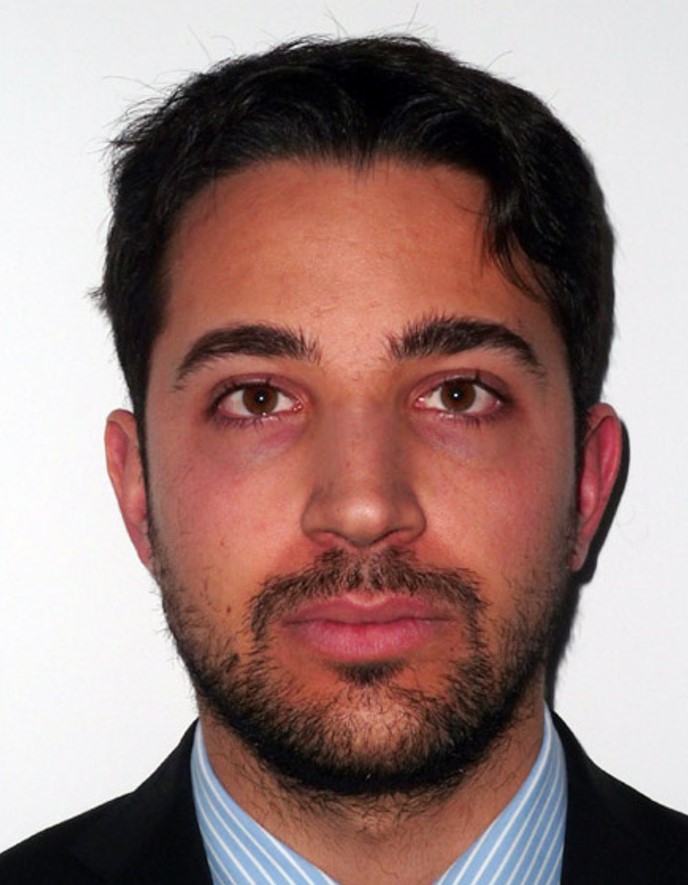}}}]{Sebastiano Fichera} received his Graduate (2010) and PhD (2014) degrees in aerospace engineering from Politecnico di Milano, Italy. He is Lecturer in aerospace engineering at the University of Liverpool, UK, where he responsible for the low-speed wind tunnel facility. Since 2020, he is also co-founder and Technical Director of the spin-off company Robotiz3d Ltd which aims to introduce autonomous systems in the road maintenance industry. His research interests lie in the field of linear/nonlinear experimental aeroelasticity, active control, morphing and his current activity is focused on studying aeroelastic systems in presence of uncertainties and probabilistic aeroelastic control. He is also interested in soft robotics, autonomous manufacturing, and mechatronics for healthcare.
\end{IEEEbiography}

\begin{IEEEbiography}[{\resizebox{1in}{!}{\includegraphics[width=1in,height=1.25in,clip]{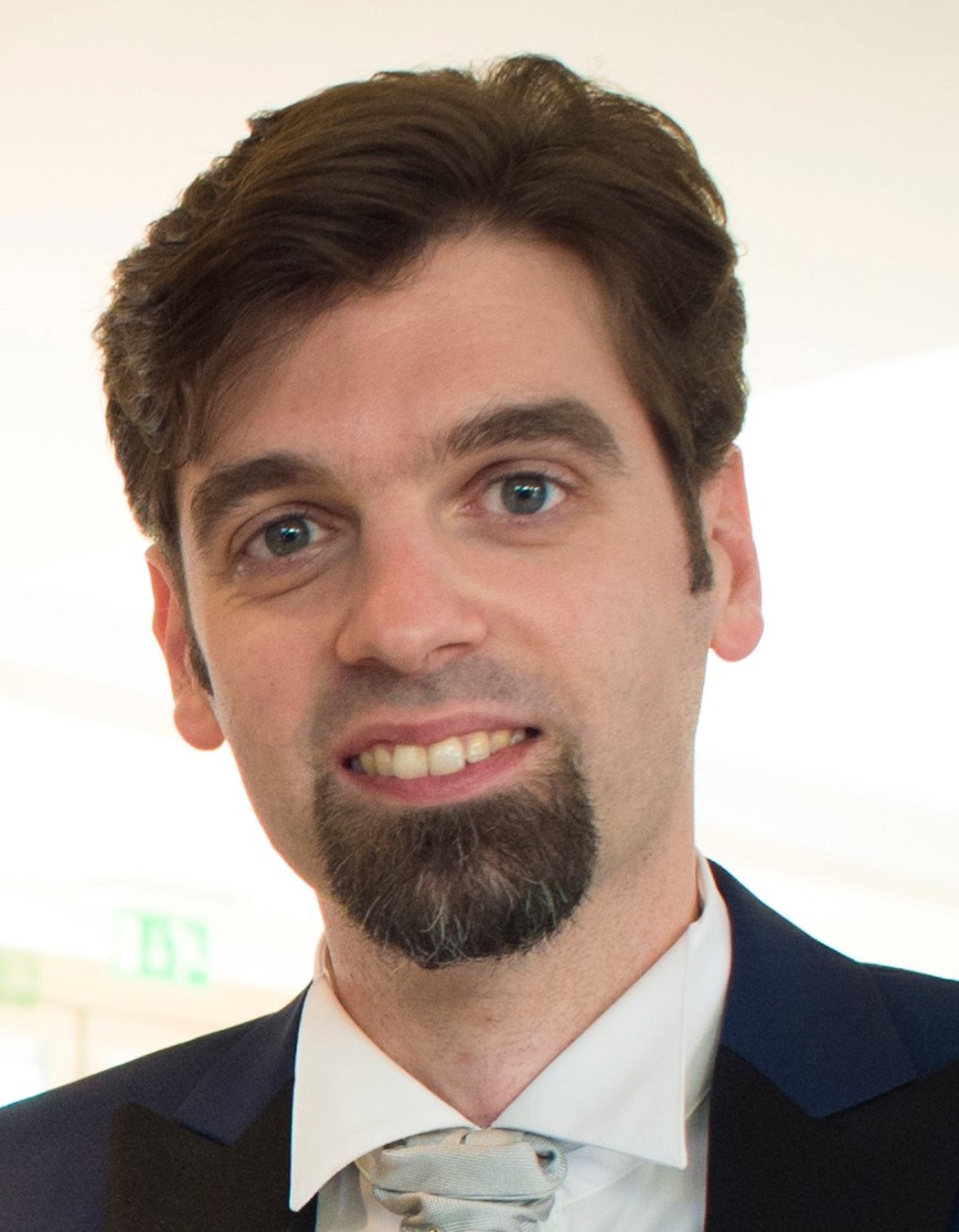}}}]{Paolo Paoletti} received the Graduate degree in automation engineering in 2006 and the PhD degree in nonlinear dynamics and complex systems in 2010 from the University of Florence, Italy. He is a Senior Lecturer in control with the University of Liverpool,  UK, where he leads the \@ Liverpool Engineering Robotics Technology (\@LERT) lab. Since 2020, he is also co-founder and CTO of the spin-off company Robotiz3d Ltd which aims to introduce autonomous systems in the road maintenance industry. Previously, he worked as an Research Assistant with the Italian Institute for Complex Systems (2010), and as a Postdoctoral Fellow with the School of Engineering and Applied Science, Harvard University (2010–2012). His research interests include nonlinear dynamics and control, with a special focus on problems that sit at the boundary between different traditional disciplines such as biology, robotics, computer science, mathematics, and physics. In 2014, he was a recipient of the “Rising Star” award from the UK Engineering and Physical Sciences Research Council.
\end{IEEEbiography}
\vspace{-7ex}

\begin{IEEEbiography}[{\resizebox{1in}{!}{\includegraphics[width=1in,height=1.25in,clip]{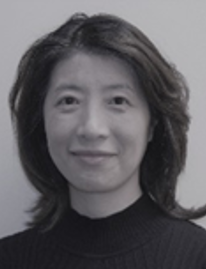}}}]{Lisa Layzell} has a BSc majoring in Genetics, and an MBA from Alliance Manchester Business School. She is an award-winning serial entrepreneur, management consultant, investment board member, and honorary professor. With a career spanning biotech, robotics, AI and cloud computing, she is experienced in commercialising technology and identifying disruptive emerging trends relevant to an organisation’s long-term innovation goals. Served as Chief Executive and led both multinational and startup companies through major revenue growth and geographical expansions. She co-founded Robotiz3d in 2020, serving as CEO, focusing on data-driven, autonomous solutions.
\end{IEEEbiography}

\begin{IEEEbiography}[{\resizebox{1in}{!}{\includegraphics[width=1in,height=1.25in,clip]{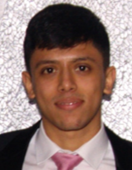}}}]{Devansh Mehta} received his MEng degree in Systems and Control Engineering from the University of Sheffield. He is currently a robotics engineer at Robotiz3d, where he is developing self-driving robotic vehicles for the road maintenance industry. Previously, he worked as a mechanical engineer, commercialising new electromechanical proteomics and microfluidics technology to analyse Bone Turnover Markers. His interests include autonomous vehicles, healthcare robotics, and AI.
\end{IEEEbiography}

\begin{IEEEbiography}[{\resizebox{1in}{!}{\includegraphics[width=1in,height=1.25in,clip]{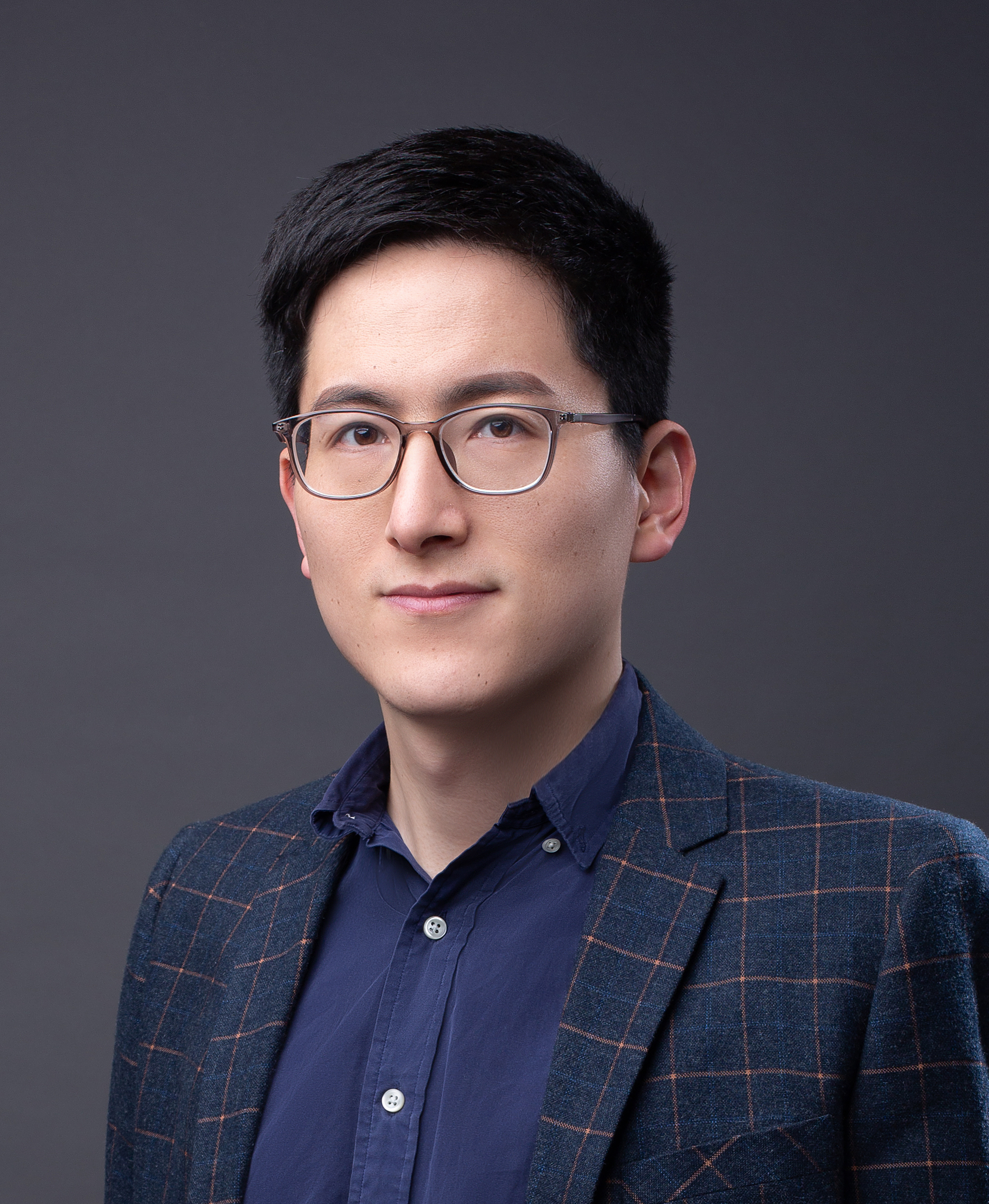}}}]{Shan Luo} received the B.Eng. degree in Automatic Control from China University of Petroleum, Qingdao, China, in 2012, and the Ph.D. degree in Robotics from King’s College London, London, U.K., in 2016. He is a Reader (Associate Professor) in Robotics and AI with the Department of Engineering, King’s College London, where he leads the Robot Perception Lab (RPL). Previously, he was a Lecturer with the University of Liverpool, and Research Fellow with Harvard University, Cambridge, MA, USA, and University of Leeds, Leeds, U.K. He was also a Visiting Scientist with the Computer Science and Artificial Intelligence Laboratory, Massachusetts Institute of Technology, Cambridge, MA, USA. His research interests include tactile sensing, robot visual perception, robot learning and robot visual-tactile perception.
\end{IEEEbiography}
\end{document}

%% file: 01_introduction.tex
% Main text
\section{Introduction}
\label{sec:intro}
Road conditions are crucial for traffic safety, economic development, and growth. Defective roads increase the likelihood of vehicle accidents and pose significant threats to traffic safety. In the United Kingdom, for example, poor or damaged road surfaces were the cause of 12.6\% of all car accidents in 2020, according to the Department for Transport~\cite{uk2022caraccident}. As autonomous vehicle driving becomes increasingly prevalent, it is essential to address road surface defects to ensure safe operations and enable fully autonomous deployment~\cite{lim2011developing,prasanna2014automated}.

Road defects also have negative environmental consequences. Research suggests that the condition of road surfaces has a significant impact on carbon emissions, with higher emissions being observed when road surfaces are in poor condition~\cite{Alsaadi2021roadcarbon}. Furthermore, a study conducted in~\cite{Wang2020roadcarbon} demonstrated that poor road surface quality leads to a 2\% increase in carbon emissions and a 5\% increase in fuel consumption. 

Monitoring road conditions is crucial for ensuring road safety and managing road infrastructure. The traditional practice of manual defect detection and maintenance is time-consuming and expensive and poses risks to the workers who perform maintenance tasks on the road. In recent years, there has been a significant increase in interest in developing automatic road defect detection systems to address these challenges. Thanks to the remarkable progress in the field of computer vision and machine learning, there have been rapid developments in visual road defect detection methods, including both traditional image processing techniques and deep learning methods, as extensively reported
in~\cite{chambon2011automatic, zakeri2017image, scholar2018review, cao2020review, hsieh2020machine,peel2018localisation}. 

Image-based methods are the most widely used approach for road surface defect detection, as reported in the literature \cite{salari2011pavement1,salari2011pavement2,peng2015research,wang2011pavement,zhao2010improvement,ayenu2008evaluating,douka2003crack,chen2017nb,ren2015faster,fang2020novel,mandal2018automated,jenkins2018deep,konig2019convolutional}. Typically, these methods involve extracting informative features from road surface images and then detecting defects using these features. The detection of road defects can be performed using various methodologies, such as image segmentation \cite{jenkins2018deep,zou2018deepcrack,konig2019convolutional,ji2020integrated,chen2018encoder,zhang2020crackgan} or object detection \cite{chen2017nb,liu2019deepcrack,fang2020novel,mandal2018automated}. In these approaches, it is generally assumed that the pixel intensities or texture information of defects differ from normal road surfaces \cite{salari2011pavement1,salari2011pavement2,liu2019deepcrack}.

Image-based methods dominate the literature on road defect detection, and several review articles have been published to survey these methods. For instance, in~\cite{chambon2011automatic}, image processing techniques for road crack detection are summarised. In~\cite{zakeri2017image}, a review of various platforms and image processing approaches is provided for evaluating the conditions of asphalt surfaces. In~\cite{hsieh2020machine}, researchers can find a survey of Machine Learning-based crack detection algorithms. These review articles offer valuable insight into image-based road surface defect detection, providing researchers with a better understanding of the existing methods and their strengths and weaknesses.

%Add more review paper information
However, existing surveys on road defect detection mainly focus on image-based methods and do not cover the recent trends of using non-image data for this purpose. In recent years, there has been a growing interest in utilising depth maps \cite{yu20143d,tsai2018pothole} or point clouds \cite{yang2020intelligent,du2020pothole} for road defect detection. Compared to image data, non-image data is less sensitive to changes in climate and illumination conditions, making it a more robust solution. Several methods have been proposed that utilise features obtained from non-vision sensors, such as laser scanners or LiDARs \cite{yu20143d,tsai2018pothole,yang2020intelligent,du2020pothole,kim2020automated,nasrollahi2019concrete}. In addition, data-fusion approaches that integrate features from images and other sensing modalities have been developed. Given the importance of road defect detection as an industrial application and the need to accelerate advancements in this field, a new investigation that reflects the recent research trend is needed.

In this paper, our goal is to offer an exhaustive exploration of existing techniques for road surface anomaly detection and pinpoint potential directions for upcoming research. We commence by elaborating on the diverse materials employed in road construction and the variety of associated road defects. We also delve into an array of sensors employed to gather relevant data for pinpointing these road defects and introduce the datasets that are openly accessible in the literature. To mirror the current trajectory of research, we divide the forefront of road anomaly identification into three distinct strategies: techniques rooted in imagery, those that use non-image data, and those that incorporate data fusion. Initially, we assess the advancements in methods centred around images, emphasising how factors like climatic variations and lighting conditions might influence their efficacy. We then present an overview of road defect detection techniques that use other sensing modalities, such as depth sensing, to overcome the limitations of image-based methods \cite{yu20143d,tsai2018pothole,yang2020intelligent,du2020pothole,nasrollahi2019concrete,kim2020automated,nasrollahi2019concrete}. % Laser scanners, for example, create point clouds by directly irradiating the road surface for distance measurement, making them more robust to changes in weather and natural illumination.
It is worth noting that, for the first time, academic researchers and industrial practitioners have collaborated in this survey to provide a more comprehensive understanding of this field.

This paper is organised as follows. In Section \ref{sec:2}, we present an overview of road surfacing materials and different types of road defects. Information about various sensors, such as cameras or laser scanners, is provided in Section \ref{sec:3}. Section \ref{sec:4} describes various publicly available road surface defect detection datasets. In Section \ref{sec:5}, we categorise various defect detection methods based on the input data type and their methodologies and provide detailed information on each method. Finally, we conclude this paper in Section \ref{sec:6}.

%% file: 02_Road_defects_categories.tex
\section{Road types and defects}
\label{sec:2}
Road surfaces are made up of various materials, each with its own characteristics and performance. The type of road defects that occur are also related to the material of the road. In this section, we will provide a detailed introduction to various materials used for road surfacing and discuss some representative road defect types that are mainly considered in road surface defect detection studies. 

\subsection{Road types depending on surfacing materials}
\begin{figure}[t]
	\centering
        \includegraphics[width=\columnwidth]{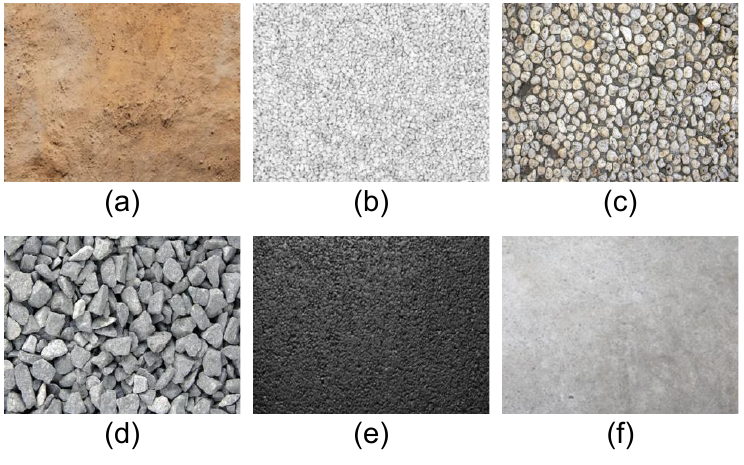}
        % \fbox{\rule{0pt}{2in} \rule{\linewidth}{0pt}}
        \caption{Images of various types of roads based on the materials used for road surfacing: (a) Earthen road, (b) Gravel road, (c) Kankar road, (d) WBM road, (e) Asphalt road, and (f) Concrete road.}
        \label{fig:Roadtypes}
	\vspace{-2ex}
\end{figure}

In today's society, there is a wide variety of road surfacing materials being used. As shown in Fig. \ref{fig:Roadtypes}, these materials include soil, gravel, Kankar, asphalt and concrete. Firstly, we will briefly introduce these road surfacing materials and compare their properties and characteristics.

An \textbf{earthen road} is typically constructed using natural soil and is considered a cost-effective option due to the abundance of soil. However, the road surface made from the soil is not durable as the physical properties of the road change with weather conditions. Earthen roads are commonly constructed in areas with low traffic, such as rural and countryside regions.

Stabilised Earthen Road is a solution that overcomes the limitations of a regular earthen road. It involves adding an additional compacted layer using stabilised soil, which can improve the road's durability and strength. Stabilised soil is created by incorporating physical, chemical, or biological agents into ordinary soil, resulting in a more robust and durable material \cite{afrin2017review}.

A \textbf{gravel road}, also known as a "metal road" or "dirty road", is a type of unpaved roads that is surfaced with gravel. The gravel used to construct the road is typically obtained from a quarry or stream bed. While these roads are common in developing countries and rural areas, they are also prevalent in many developed countries \cite{skorseth2000gravel}. In fact, according to Rajkamal \etal \cite{rajkamal2016performance}, more than 75\% of the road network in many developing countries consists of gravel and earth roads. Gravel roads are more durable than earthen roads and can withstand climate change. However, the surface of a gravel road made from natural gravel may not be uniform, damaging the lower part of vehicles.

\begin{figure*}
	\centering
        % \fbox{\rule{0pt}{2in} \rule{\linewidth}{0pt}}
        \includegraphics[width=\linewidth]{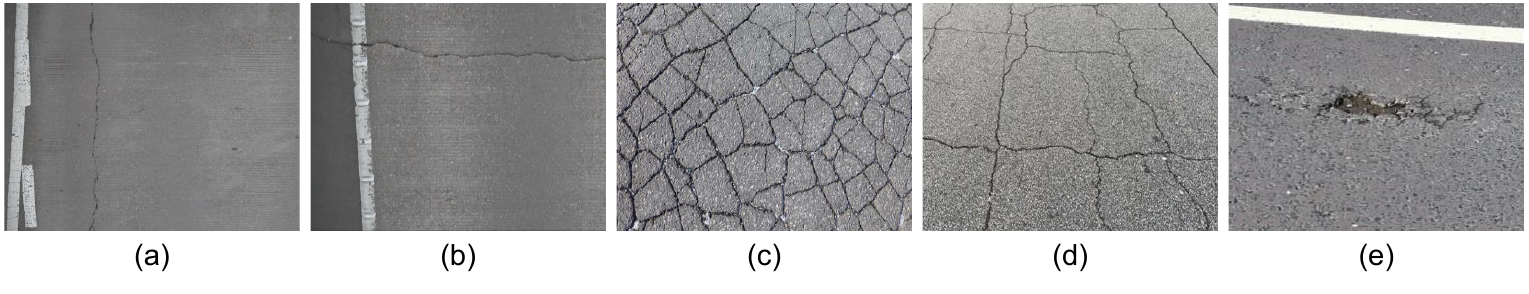}
           \caption{Example snapshots of various types of road surface defects: (a) Transverse crack, (b) Longitudinal track, (c) Alligator crack, (d) Block crack, and (e) Pothole. It is important to note that T
          the key difference between (a) and (b) is the angle with the road lane. If the crack runs parallel to the road lane, it is classified as a transverse crack. Conversely, if the crack is at a right angle to the lane, it is classified as a longitudinal crack.}
        \label{ig:defects}
	\vspace{-2ex}
\end{figure*}

A \textbf{Kankar road} is a type of road constructed using Kankar, a nodular variety of limestone that is spongy in nature and contains clayey and silicious matter. Kankar is a common surfacing material in countries with abundant limestone mines. However, the quality of the road surface depends on the quality of Kankar used. While a hard variety of Kankar may be as strong as stones, a soft variety may be weak. The construction cost of a Kankar road is similar to that of earthen and gravel roads, making it a popular choice in developing countries.

A \textbf{WBM road}, short for Water Bound Macadam road, is constructed with a stone border. The border of a WBM road is filled with sand, soil and very fine gravel called "Screening", which fastens the stone borders. The WBM road is more robust in the rain compared with an earthen road since the top-wearing surface is made up of stone gravel and a boundary. Moreover, it provides better functionality than graver roads and Kankar roads that are constructed using single materials, as the soil helps to improve the flatness of the road. In the construction of the WBM road, a rolling machine is used to compact the stone border, while water is added to it during the rolling process. This kind of road is particularly effective and compacted in locations with heavy rainfall.

In the case of roads constructed with various pavement materials, as mentioned earlier, there is a lack of homogeneity in the pavement structure. For instance, gravel roads exhibit unevenness due to variations in gravel particle sizes. While these roads are not prone to surface cracks such as longitudinal or alligator cracks, they may develop potholes over time due to the sinking of the road caused by vehicle weight-induced pressure. Detecting such defects can be conveniently accomplished by utilising laser-based measurements to determine the distance of the road surface, as highlighted in a study by Pan \etal~\cite{pan2018detection}.

An \textbf{asphalt road} is a road surfaced with asphalt. Asphalt is a material that is produced through the refining of petroleum. To improve its durability and strength for road surfacing, it is mixed with aggregate, bitumen and sand and then heated and dried. This is then installed across a surface, such as an asphalt driveway, to form an asphalt road. Compared with the aforementioned road types, asphalt roads provide better functionality due to their smooth surface that boosts traction and skid resistance. Additionally, asphalt roads have an attractive price point due to their 100\% recovery rate. Therefore, asphalt is the primary material used to build heavy-traffic roads. 

Most major roads in areas with high traffic are made of asphalt, which can make managing them expensive \cite{shahin2005pavement}. In fact, Highway Statistics \cite{statistics2016federal} report that in 2016, approximately 94\% of asphalt surfaces were made of asphalt. As a result, advanced maintenance technology is available for asphalt roads, unlike other roads such as earthen, gravel, or Kankar roads. This is reflected in the majority of studies which propose detection techniques specifically for defects occurring on asphalt roads.

A \textbf{concrete road} is made by mixing aggregates such as crushed rock, sand, cement, and water. The cement serves as the binder, holding the aggregate together as it dries into a hard, brittle surface that can crack and break, especially if the underlying surface is uneven. To address this issue, a mixture of cement and stone gravel is used when constructing concrete roads, making them more resistant to wear and tear problems like rutting, cracking, stripping loss of texture, and potholes commonly found in flexible pavement surfaces.

Concrete road paving has become increasingly popular in recent years, rivalling asphalt paving in popularity. This trend is mainly due to concrete roads' superior durability, longevity, and weather resistance. Unlike asphalt roads, which last for around 10 years, concrete roads have a lifespan of over 20 years due to their robustness. Additionally, concrete roads have low thermal expansion and contraction properties, which prevent surface defects from developing \cite{wang2012analysis}. However, a major challenge with concrete roads is their repair, which requires the replacement of entire slabs rather than simple patching. Furthermore, the installation and maintenance costs for concrete roads are higher than those for asphalt.

\subsection{Types of road defects}
Road surfaces can develop various types of defects, including potholes, surface deterioration, edge failure, cracking, crazing, rutting, and subsidence, as illustrated in Fig. \ref{ig:defects}. These defects can arise due to various factors, such as climate change, vehicle accidents, and the natural ageing of road surfaces. In this survey paper, we provide detailed information on the typical patterns of road surface defects. 

\textbf{Longitudinal and transverse cracks} are two common types of cracks found inroads. Longitudinal cracks run parallel to the road centerline, while transverse cracks run perpendicular to the traffic direction. Transverse cracks typically appear on both sides of the road and may eventually develop into horizontal cracks through the road. In some cases, these cracks may be accompanied by a small number of branching cracks. 

A variety of factors can cause longitudinal cracks. Firstly, improper joint placement or construction can be a cause. This is because joints are often challenging to compact fully, resulting in a weak spot in the pavement. Secondly, a reflecting crack from a sublayer (not including joint reflection cracking) can cause a crack to develop directly on top of an underlying fracture due to minor movement of the underlying crack that stresses the overlying layer. Furthermore, fatigue caused by Hot-Mix Asphalt is another significant factor contributing to longitudinal cracks. Initially, signs of fatigue cracking typically manifest in the longitudinal direction, either directly within or near the wheel path. In pavements with considerable thickness relative to the applied load, cracking can initiate from the pavement surface instead of originating at the bottom of the pavement layer. This occurrence is attributed to surface-induced stress generated by wheel loads, which, when combined with an aged and more brittle asphalt surface layer, leads to the formation of cracks.

\textbf{Alligator crack}, also known as crocodile crack, is a type of pavement distress that results in the formation of small acute polygons resembling a turtle pattern. These cracks are caused by repeated traffic loads that lead to longitudinal fissures along the wheel paths. As these fissures continue to develop, they eventually form a distinctive pattern resembling alligator skin. This type of pavement distress is typically due to a decrease in the base, sub-base, or subgrade support's capacity to support loads as a result of poor drainage or spring thaw problems. Other contributing factors may include overload, inadequate pavement structural design, insufficient thickness to support the intended loads, poor compaction and subgrade preparation during pavement construction, and improper pavement drainage design.

\textbf{Block crack} is characterised by interconnected cracks that divide the pavement into roughly rectangular pieces. In some cases, block cracking results from longitudinal and transverse cracks that break the pavement into chunks of various sizes. These cracks can be as small as a foot or as large as several yards square. As the cracks deepen and spread, moisture and vegetation growth further degrade the surface, rendering it unsuitable for foot traffic or vehicle use. The shrinkage of the surface material and temperature changes over time primarily causes block cracking. Proper mixing and placement of the asphalt mix, as well as the use of suitable asphalt binders, can help prevent the occurrence of block cracks.

A \textbf{pothole} is a sharply edged hole that forms on the upper layers of a road surface, typically due to untreated alligator cracks and moisture ingress. As alligator cracking worsens, small pavement sections become loose and can move under the weight of passing vehicles, leaving a gap in the surface layer that forms a pothole. Potholes may seem minor, but they can lead to serious consequences, including billions of dollars in vehicle damages and a significant proportion of highway fatalities. Unlike other types of road damage that have a negligible impact on driving, potholes can cause temporary loss of traction and lead to accidents. It is important to repair potholes promptly to prevent further deterioration of road surfaces and maintain safety for road users. According to the American Automobile Association, around 16 million drivers in the United States experienced vehicle damage caused by potholes in the five years leading up to 2016, resulting in punctured tyres, bent wheels, and damaged suspensions, with an annual cost of \$3 billion \cite{american2016pothole}. In the United Kingdom alone, repairing all potholes on the country's roads is estimated to cost £12 billion \cite{song2018pothole}.

\section{Data collection protocols} 
\label{sec:3}
There are various types of sensors used to detect road surface defect detection, with cameras being the most commonly used. However, the image captured by cameras is susceptible to changes in climate and illumination conditions. Recent studies have proposed using depth sensors based on Time of Flight (ToF) sensors or laser scanning technology, such as LiDAR, to overcome this limitation. This section introduces different types of sensors used in road defect studies. We cover the sensors used in research and describe off-the-shelf platforms that utilise sensors to collect road surface information for defect detection.

\subsection{RGB Camera}
RGB cameras are widely used for detecting road surface defects, with many researchers and companies in the road management and defect-fixing fields relying on them to capture images of road surfaces \cite{jenkins2018deep,zou2018deepcrack,konig2019convolutional,ji2020integrated,chen2018encoder,zhang2020crackgan}. From simple mobile cameras mounted on cell phones \cite{zhang2016road,shi2016automatic} to ultra-high-resolution cameras of 4K or higher \cite{eisenbach2017get,zou2012cracktree,konig2021optimized,liu2019deepcrack}, various types of RBG cameras have been used in both research and commercial products for road defect detection. Typically, methods for detecting road surface defects using RGB images aim to identify differences in texture and pixel intensities between the normal road surface and those with defects. Several studies have utilised RGB cameras for road surface defect detection, including \cite{jenkins2018deep,konig2021optimized}.

Mobile cameras have become popular for collecting road surface images due to their affordability and accessibility. Since almost everyone owns a mobile device, the biggest advantage is that it does not require additional costs to purchase a camera. Mei and Gul presented a deep neural network for detecting pavement cracks, using the camera module on an iPhone 5 to collect the road surface images \cite{mei2020cost}. The Crack500 \cite{zhang2016road} and CrackForest \cite{shi2016automatic} datasets, commonly used for road defect segmentation, are composed of images taken using cell phones. However, the image sensor scale of mobile cameras is relatively small compared to general digital cameras, resulting in lower-resolution images. For example, an image in the CrackForest dataset has a resolution of 480 $\times$ 320. 

General digital cameras are another popular tool for collecting road surface data due to their relatively high resolution. Hu \etal used a Nikon digital camera to capture over 3,000 road surface pictures with a resolution of 2976 $\times$ 3968 \cite{hu2021pavement}. Similarly, Yusof \etal used a digital camera to generate road surface image samples for training a convolutional neural network that detects cracks on asphalt pavement \cite{yusof2019deep}. These cameras are useful in research and installed in off-the-shelf platforms for road defect detection.

\begin{figure}[t]
	\centering
        \includegraphics[width=\columnwidth]{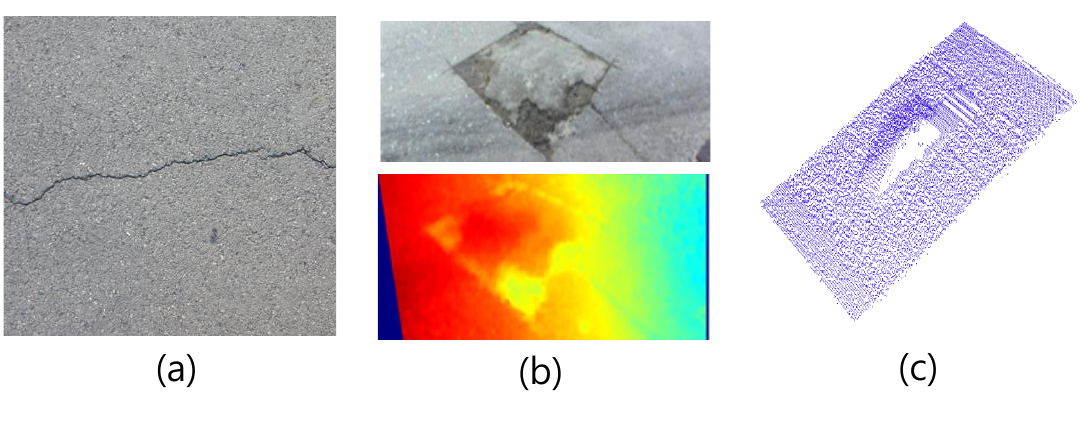}
        % \fbox{\rule{0pt}{2in} \rule{\linewidth}{0pt}}
        \caption{Example snapshots of data collected from various sensors, including RGB cameras, stereo cameras, and Time of Flight (ToF) based depth sensors: (a) An image from the CrackForest dataset \cite{shi2016automatic} taken by an iPhone 5's mobile camera. (b) An image of a road surface and the corresponding disparity map obtained by Fan \etal \cite{fan2018real}. (c) Visualisation results of point clouds collected by Zhu \etal \cite{zhu2020measurement}.}
        \label{fig:data_format}	
 \vspace{-2ex}
\end{figure}

One of the main advantages of using image sensors for road defect detection is their availability and the wealth of existing studies, which facilitates the application of various models. However, as discussed earlier, image-based road defect detection methods rely on detecting defects by analysing the difference between the texture and pixel values of the normal road surface and the defect. As shown in Fig.~\ref{fig:data_format}(a), road defects typically appear darker than normal road surfaces due to their deeper nature. These challenges can compromise the robustness of image-based methods in certain scenarios, including lighting conditions (e.g., day versus night) and alterations in road surface texture resulting from weather changes. Moreover, changes in road materials or the presence of reflective surfaces can also impact the reliability of information captured by image sensors. Addressing these challenges often necessitates a substantial volume of image data, which can present a significant obstacle. This is why various explicit depth sensors, such as ToF-based or structural light-based sensors, are starting to become an alternative to cameras. However, visual cameras are still a predominant way to gather information for road defect detection since they are low-cost and easy to access. Numerous image-based road defect detection methods have been proposed and will be elaborated on later.

\subsection{Stereo cameras}
Numerous studies, including those by Li \etal~\cite{li2018road}, Zhang \etal~\cite{zhang2014efficient}, Thompson \etal~\cite{thompson2022shrec}, and Fan \etal~\cite{fan2019pothole}, have demonstrated the feasibility of robust road defect detection using 3D methods. One commonly employed approach to obtain 3D information is through the use of a stereo camera, which captures images of the same subject from multiple angles and utilises the disparity that is the distance between two sets of coordinates of the cameras for the same scene point. This method offers a cost advantage as it allows for the estimation of 3D information without needing an additional sensor like a laser scanner to acquire depth information. Li \etal~\cite{li2018road} employed a stereo camera consisting of two USB cameras for pothole detection, while Zhang \etal. Utilised a stereo camera constructed using two PointGrey Flea 3 cameras. Figure~\ref{fig:data_format}(b) presents an example of data collected from a stereo camera.

While using a stereo camera to obtain 3D information has its advantages, there are also drawbacks. One limitation is the difficulty in handling textureless or reflective surfaces. When a surface lacks distinctive features, the stereo-matching algorithm may struggle to find reliable correspondences, leading to inaccurate depth estimation in such areas.

Furthermore, deploying a practical stereo camera system requires manual calibration. Precise calibration is essential to ensure accurate depth estimation. Calibration involves determining the cameras' intrinsic parameters (focal length, distortion) and extrinsic parameters (relative position and orientation). Calibration errors can introduce inaccuracies into the depth maps, making the calibration process critical but sometimes time-consuming.

Despite these disadvantages, stereo cameras remain a popular choice for depth estimation due to their affordability, simplicity, and effectiveness within their limitations. However, it is important to consider these challenges and potential limitations when using stereo cameras for 3D information acquisition.

\subsection{ToF-based and Structured light-based depth sensors}
Several road defect detection approaches using depth sensors based on Time-of-Flight (ToF) and structured light-based methods have been introduced in recent years. Unlike stereoscopic cameras, which estimate depth based on the disparity of images from two cameras, these sensors measure distance directly by projecting laser or structured infrared rays onto the road surface. This makes them more robust to external environmental conditions and eliminates the need for heuristic parameters for camera calibration.

Among the structured light sensors commonly used for detecting road defects, one prevalent depth-sensing device is the Microsoft Kinect \cite{sarbolandi2015kinect}. The Kinect utilises infrared sensors and has an approximate depth measurement accuracy of $\pm$ 5\% \cite{jahanshahi2013unsupervised,wang2017asphalt,thompson2022shrec}. Another popular choice is the RealSense L515\footnote{\url{https://www.intelrealsense.com/lidar-camera-l515/}}, a structured-light-based depth sensor extensively used for road defect detection \cite{gu2022surface,kumar2016cost}.

Compared to depth sensors based on stereo cameras, structured-light-based depth sensors such as the Kinect and RealSense L515 offer accurate depth measurements of the road surface and defects. This high level of accuracy is crucial for effectively detecting and characterizing road defects.

LiDARs have been used to detect road defects \cite{chen2022gocomfort,fan2019pothole,zhang2019intelligent} among the ToF-based depth sensors. Similar to the structured-light-based depth sensors, these sensors measure the explicit distance between the sensor and the road surface, making them a more straightforward way to detect road damage compared to other data collection approaches. Data collected from LiDARs is represented as point clouds, which are discrete sets of data points in space, as shown in Fig.~\ref{fig:data_format}(c). This format enables the structural form of road defects to be expressed in three dimensions, in contrast to other data formats that can only capture the external form of road defects.

However, not all LiDARs are suitable for road surface assessment. For near-field perception, some LiDARs have a long perception range that may not be effective. Several commercial LiDARs are specialised for road assessment, such as the Velarray M1600\footnote{\url{https://velodynelidar.com/products/velarray-m1600/}}. The XenoTrack\footnote{\url{https://xenomatix.com/lidar/xenotrack/}} and Trimble X7 Scanner\footnote{\url{https://geospatial.trimble.com/x7-scanner}} are two other commercial LiDARs designed for road assessment. Trimble Geospatial\footnote{\url{https://geospatial.trimble.com/}} offers the high-speed 3D laser scanning system, Trimble X7 Scanner, which simplifies adoption and increases efficiency, as well as complete field-to-finish mobile mapping solutions, such as Trimble MX9\footnote{\url{https://geospatial.trimble.com/products-and-solutions/mobile-mapping}} and Trimble MX50. Compared to Trimble MX9, Trimble MX50 has a slightly higher laser scan accuracy (2mm) and precision (2.5mm). Velodyne's M1600 provides near-field perception up to 30 meters away and a wide vertical field of vision of 32 degrees. The L515 is a LiDAR depth camera that consistently delivers high accuracy across a range of 0.20m to 9m and provides over 23 million precise depth pixels per second, with a depth resolution of 1024 x 768 at 30 frames per second. On the other hand, the XenoTrack LiDAR system from Xenomatix\footnote{\url{https://xenomatix.com/}} is designed for road assessment and has high precision and high-resolution short-range capabilities that enable granular surface measurements and reveal intricate road geometries. The XenoTrack is mountable on vehicles and can be used instantly without complex setup procedures.

There are several off-the-shelf laser systems available for crack measurement, and one of the most popular is the Laser Crack Measurement System (LCMS®-2)~\footnote{\url{https://www.pavemetrics.com/applications/road-inspection/lcms2-en/}} from
Pavemetrics~\footnote{\url{https://www.pavemetrics.com/}}. LCMS-2 is a single-pass 3D sensor for pavement inspection that can be used to automatically geo-tag, measure, detect and quantify all key functional parameters of pavement. It provides a complete 1mm resolution automated pavement condition survey and can dramatically reduce labour costs.

% \begin{figure}
%     \centering
%         \caption{The LCMS Laser System from Pavemetrics.}
%         \label{fig::LaserSystem}
%         \includegraphics[width=\linewidth]{Figures/LaserSystem.png}
%         % \fbox{\rule{0pt}{2in} \rule{\linewidth}{0pt}}
% \end{figure}
% Localization sensors

% \textbf{Localization sensors.} Localization sensors such as GPS and Odometer are also important for road assessment, which can provide the pavement repairers with accurate defect locations. Trimble Geospatial provides a set of Integrated GNSS Systems that can achieve max 8 mm horizontal and 15 mm vertical localization precision\footnote{\url{https://geospatial.trimble.com/portfolios/230/compare}}.

%% file: 025_Dataset.tex
\section{Road defect detection datasets}
\label{sec:4}
In the early days of the research on road surface defect detection, datasets were not a major concern as most detection methods were based on image processing and did not require learning algorithms. However, with the rise of data-driven approaches using machine learning and deep learning, the importance of datasets has grown rapidly. While most publicly available datasets are still image-based, the use of depth sensors in research has resulted in an increasing number of datasets containing depth information. In this survey paper, we focus only on publicly accessible datasets, as private datasets are not accessible. This section provides an overview of various publicly available datasets for road defect detection. 

\subsection{RGB image-based datasets}
Image-based datasets can be categorised into two types based on their annotation methods. The first type provides pixel-level annotations for road defects, which gives pixel-wise information to distinguish between normal road surfaces and defects. These datasets usually consist of road surface images and corresponding mask images that define whether a pixel belongs to a defect or not. The second type provides bounding-box annotations, which do not provide detailed information at the pixel level. However, compared to pixel-level annotations, which can only indicate two classes (i.e., general roads and defects), bounding-box annotations can be used to classify specific types of defects by comprehensively collecting information contained in the bounding boxes.

\subsubsection{Pixel-level annotated datasets}
Over the last decade, various pixel-level annotated datasets have been made available for pavement distress detection. The CrackForest dataset \cite{shi2016automatic} is one such dataset, which comprises of 118 images of cracks on urban road surfaces in Beijing. These images were captured using an iPhone 5, resized to 480$\times$320 pixels, and labelled with pixel-wise masks. The CRACK500 dataset \cite{zhang2016road} contains 250, 50, and 200 images for training, validation, and testing, respectively, and is further divided into 12 cropped parts for each image to increase the dataset size. This results in a dataset of 1,896, 328, and 1,124 images for training, validation, and testing, respectively. The German Asphalt Pavement Distress (GAPs) dataset \cite{eisenbach2017get} is a high-quality dataset that addresses the comparability issue in pavement distress detection. This standardised dataset contains 1,969 grayscale images with various classes of distress, such as cracks, potholes, and inlaid patches. The images have a resolution of 1920 $\times$ 1080 pixels with a per-pixel resolution of 1.2 mm $\times$ 1.2 mm. The Cracktree260 dataset is an extension of the dataset used in \cite{zou2012cracktree} and consists of 260 road pavement images captured by an area-array camera. The images are rotated with nine angles (from 0 to 90 degrees, at intervals of 10), flipped vertically and horizontally for each angle, and cropped five times on each flipped image (four in the corners and one in the centre). The CRKWH100 dataset \cite{liu2019deepcrack} contains 100 road pavement images captured by a line-array camera with a ground sampling distance of 1 millimetre.

The NHA12D \cite{huang2022nha12d} dataset was collected by National Highways Surveying vehicles and consists of 80 pavement images, 40 of which depict concrete pavement and 40 depict asphalt pavement. Each image has a resolution of 1920$\times$1080 and captures the view of a single lane on the A12, with some images including additional objects such as vehicles, road markings, road studs, and water stains. Concrete pavements have transversal and longitudinal joints every certain meter, which can pose a challenge for crack detection models to differentiate. For different types of pavements, 25 images were taken using cameras with a vertical viewpoint, and the other 15 were taken with a forwarding camera. 

The EdmCrack600 dataset \cite{mei2020densely} is widely considered one of the most challenging datasets due to several factors taken into consideration during the data collection. These include changes in weather conditions, significant environmental effects, and noise such as shadows, occlusion, stains, texture differences, and low contrast due to overexposure. Furthermore, blurring effects caused by the moving vehicle and poor lighting conditions add to the complexity of the dataset.

The CQU-BPDD dataset \cite{tang2021iteratively} consists of 60,056 bituminous pavement images captured automatically by in-vehicle cameras of a professional pavement inspection vehicle from various locations in southern China at different times. Each image corresponds to a 2 × 3 metres pavement patch of highways and has a resolution of 1200×900 pixels. The dataset contains seven different distress types, including transverse crack, massive crack, alligator crack, crack pouring, longitudinal crack, ravelling, repair, and normal pavement. Due to the large number of images, it is not feasible to provide pixel-level annotation. Therefore, the CQU-BPDD dataset provides patch-level annotation. Although the dataset does not provide precise pixel-wise annotation for defining defect locations, it does provide a naive location based on a small window or patch. The dataset also includes some challenging conditions, such as uneven illumination, which is manifested as weak light in some areas, and a variety of backgrounds in natural environments, such as crossings, foreign bodies, and ruts. Table \ref{tbl:pixel_dataset} summarises the pixel-wise annotated datasets mentioned earlier in this paper.

\begin{table*}
\caption{Publicly available datasets for road surface defect segmentation.}
\label{tbl:pixel_dataset}
\centering
\setlength{\leftmargini}{0.4cm}
\resizebox{\textwidth}{!} 
{ 
\begin{tabular}{c|c|c|c|c|c|m{4cm}|c }
\toprule
\toprule
Name  & Format & Sensor  & Resolution & Label type & Surface & Description & Sample \\
\midrule
   \makecell{CrackForest\\(CFD)\\ \cite{shi2016automatic}}   &   RGB    &  \makecell{Mobile cam  \\(iPhone 5) }       &  480$\times$320        &  \makecell{Pixel-wise\\(Binary class) }        &  Asphalt   & \begin{itemize}
       \item 118 images
       \item Contain noises such as shadows, oil spots, and water stains
   \end{itemize}   & \adjustbox{valign=c}{\includegraphics[width=20mm, height=20mm]{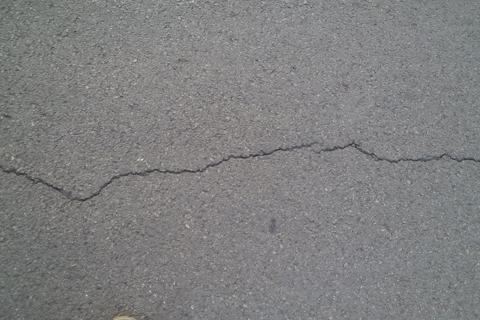}}  \\ \midrule
  
       \makecell{Crack500\\ \cite{zhang2016road} }   &   RGB    &  \makecell{Mobile cam }       &  2000$\times$1500       &  \makecell{Pixel-wise\\(Binary class) }        &  Asphalt   & \begin{itemize}
       \item 250 images
   \end{itemize}   & \adjustbox{valign=c}{\includegraphics[width=20mm, height=20mm]{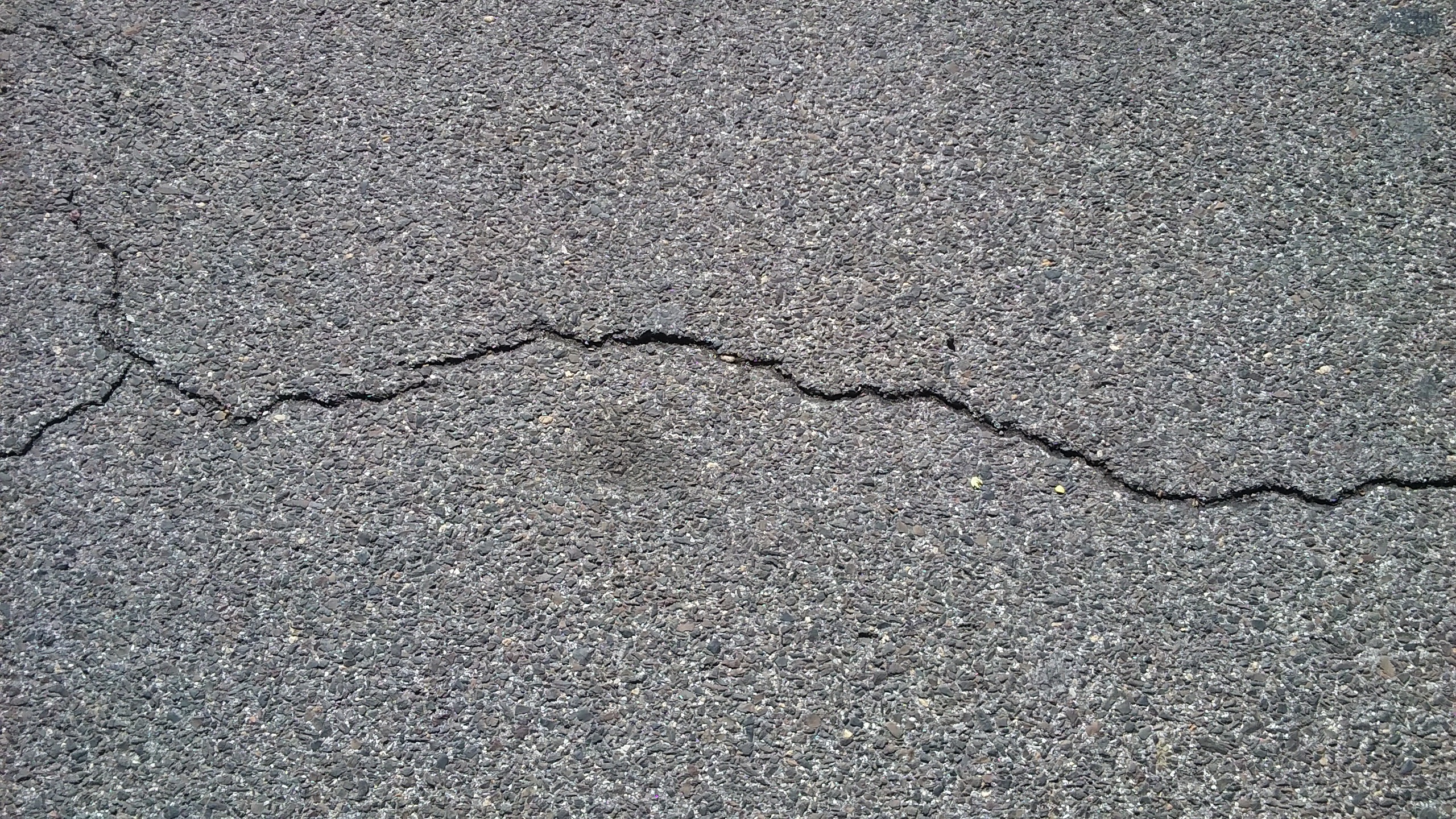}}  \\ \midrule
                  
      \makecell{German Asphalt\\Pavement Distress\\(GAPs)\\\cite{eisenbach2017get}  }   &   RGB    &  \makecell{high-resolution\\cameras}       &  1920$\times$1080        &  \makecell{Pixel-wise\\(Binary class) }        &  Asphalt   & \begin{itemize}
       \item 1,969 grey valued images
       \item various classes of distress such as cracks, potholes, inlaid patches
   \end{itemize}   & \adjustbox{valign=c}{\includegraphics[width=20mm, height=20mm]{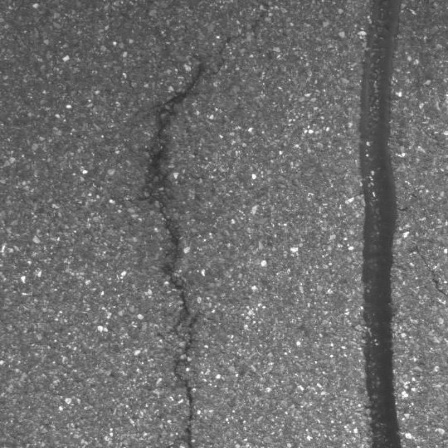}}  \\ \midrule
                  
      \makecell{CrackTree260\\ \cite{zou2012cracktree}}   &   RGB    &  \makecell{Area-array\\camera  }       &  800$\times$500       &  \makecell{Pixel-wise\\(Binary class) }        &  Asphalt   & \begin{itemize}
       \item 260 images
       \item the crack is defined byrotate the images with 9 different angles
   \end{itemize}   & \adjustbox{valign=c}{\includegraphics[width=20mm, height=20mm]{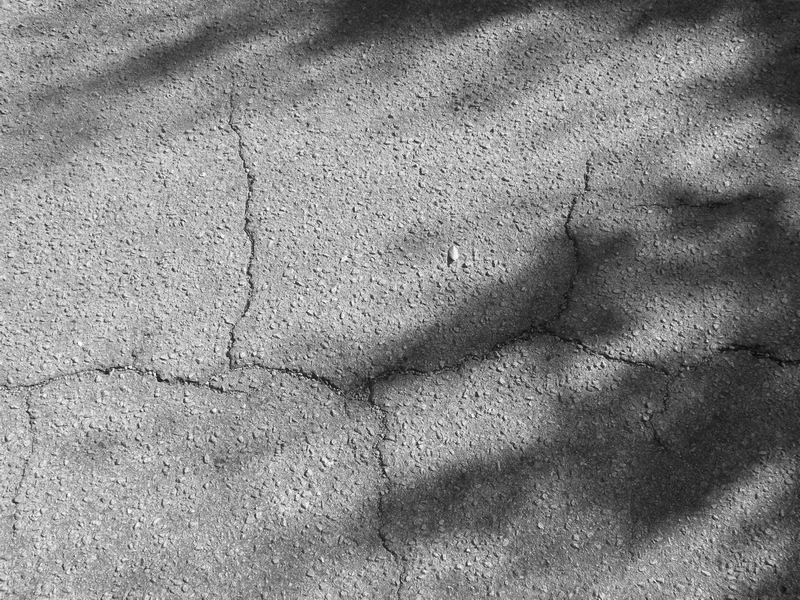}}  \\ \midrule
   
     \makecell{CRKWH100\\ \cite{liu2019deepcrack}}   &   RGB    &  \makecell{Line-array\\camera  }       &  512$\times$512       &  \makecell{Pixel-wise\\(Binary class) }        &  Asphalt   & \begin{itemize}
       \item 100 road pavement images
       \item Images captured by a line-array camera under visible-light illumination. 
   \end{itemize}   & \adjustbox{valign=c}{\includegraphics[width=20mm, height=20mm]{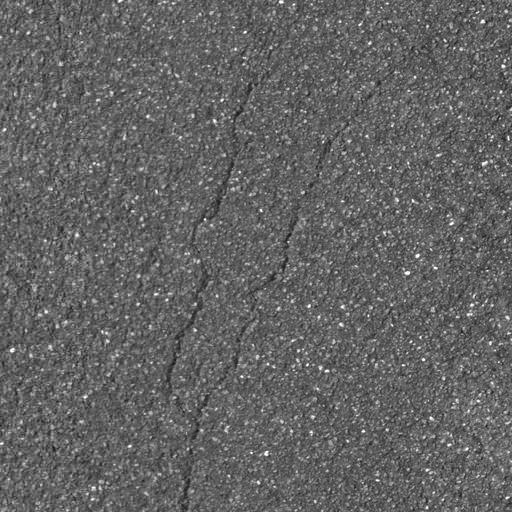}}  \\ \midrule

         \makecell{NHA12D\\ \cite{huang2022nha12d}}   &   RGB    &  \makecell{High-resolution\\camera  }       &  1920$\times$1080       &  \makecell{Pixel-wise\\(Binary class) }        &  \makecell{Asphalt\\Concrete  }    & \begin{itemize}
       \item 80 pavement images
       \item 40 concrete pavement images and 40 asphalt pavement images
   \end{itemize}   & \adjustbox{valign=c}{\includegraphics[width=20mm, height=20mm]{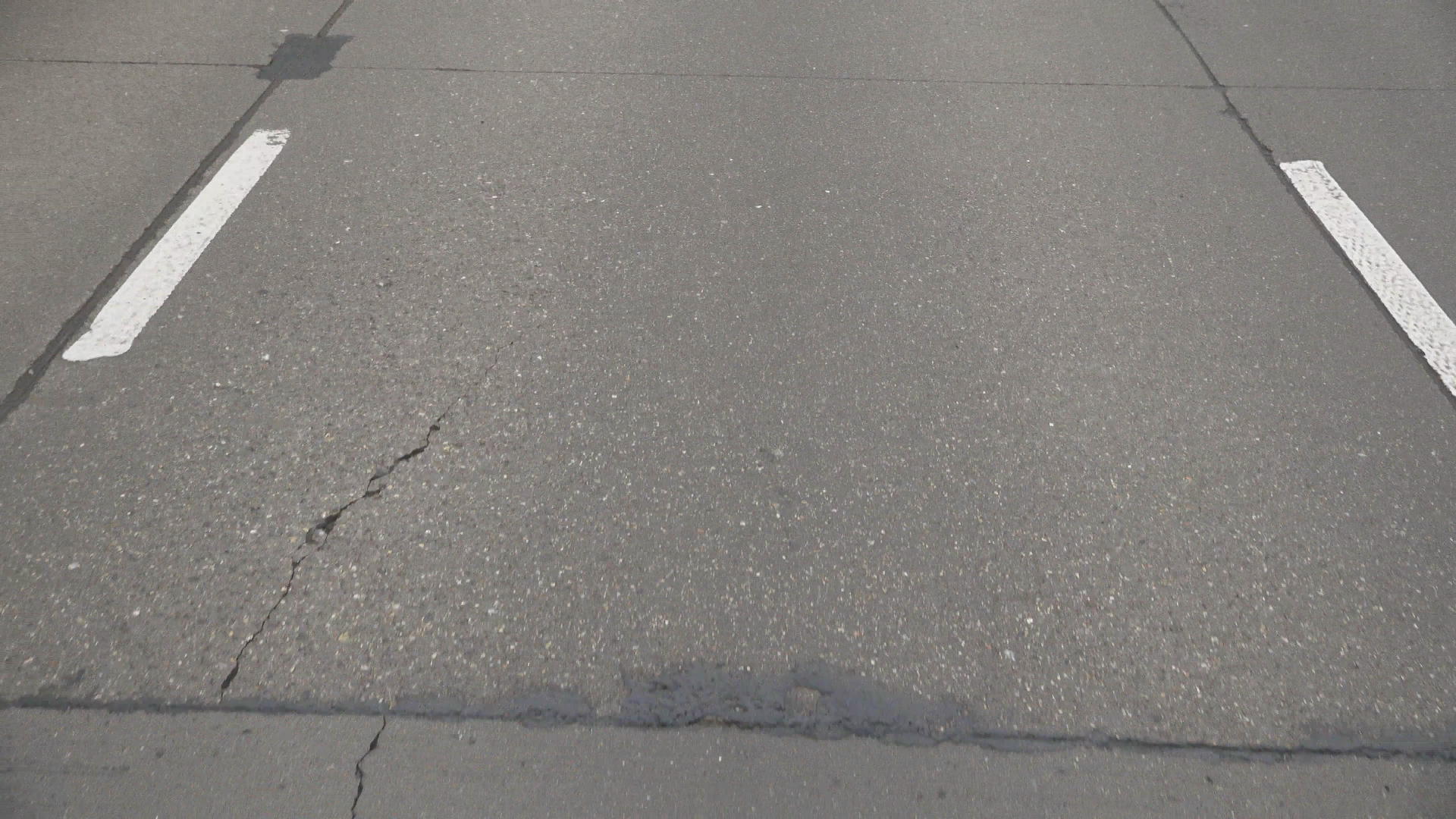}}  \\ \midrule

      \makecell{EdmCrack600\\ \cite{mei2020densely}}   &   RGB    &  \makecell{GoPro 7 }       &  1920$\times$1080       &  \makecell{Pixel-wise\\(Binary class) }        &  \makecell{Asphalt}    & \begin{itemize}
       \item 600 pavement images
       \item The data aim to cover various factors such as weather conditions, illumination conditions, and shadows from other objects.
   \end{itemize}   & \adjustbox{valign=c}{\includegraphics[width=20mm, height=20mm]{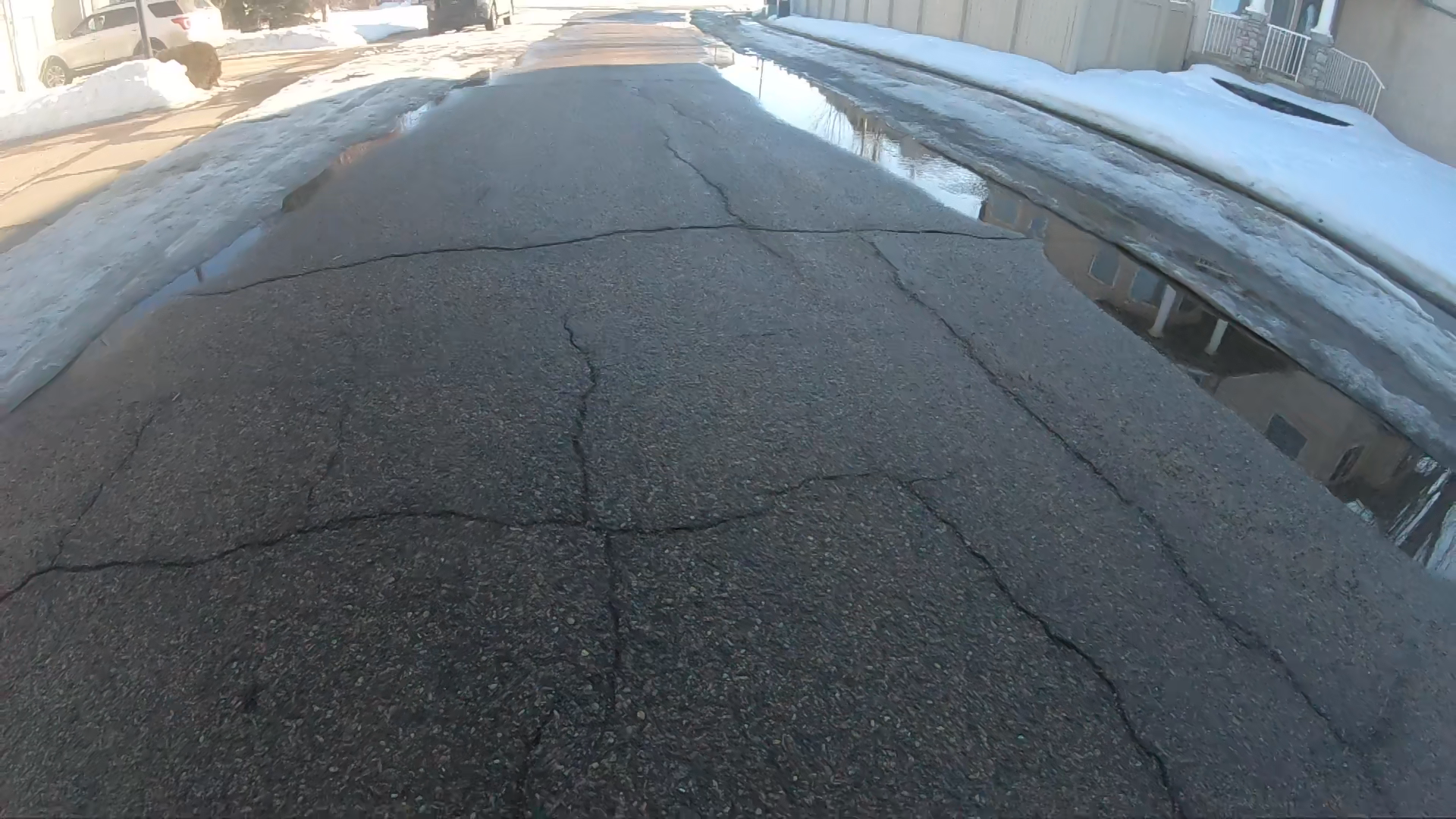}}  \\ \midrule
   
         \makecell{CQU-BPDD\\ \cite{tang2021iteratively}}   &   RGB    &  \makecell{In-vehicle\\cameras  }       &  1200$\times$900      &  \makecell{Patch-label\\(Binary class) }        &  \makecell{Asphalt}    & \begin{itemize}
       \item 60,056 pavement images
       \item Illumination is uneven, which is manifested as weak light in some areas
       \item Consider a variety of backgrounds in real environments, such as crossings, foreign bodies, ruts, etc.
   \end{itemize}   & \adjustbox{valign=c}{\includegraphics[width=20mm, height=20mm]{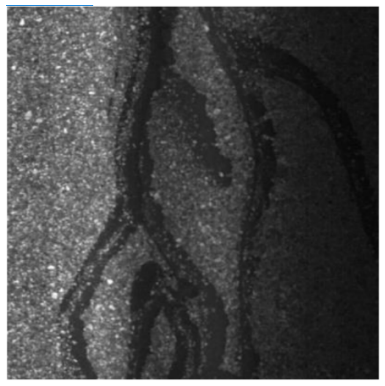}}  \\ \midrule
   
    \makecell{Stone331 \\ \cite{konig2021optimized}}   &   RGB    &  \makecell{Area-array\\camera  }       &  1024$\times$1024       &  \makecell{Pixel-wise\\(Binary class) }        &  Stone   & \begin{itemize}
       \item 331 images of stone surfaces
       \item detect stone crack observed in a cutting surface on a stone.
   \end{itemize}   & \adjustbox{valign=c}{\includegraphics[width=20mm, height=20mm]{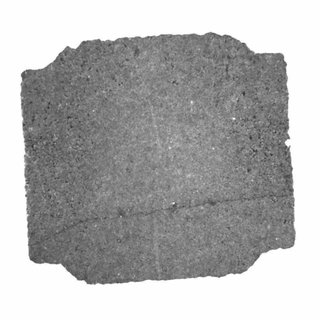}}  \\

\bottomrule
\bottomrule
\end{tabular}
}
\vspace{-2ex}
\end{table*}
\textbf{}

\subsubsection{Bounding-box annotated datasets}
While pixel-wise annotated datasets provide a variety of data, datasets in the form of bounding boxes are still limited in number. Bounding boxes offer approximate locations of cracks, with a wider field of view compared to pixel-level-annotated datasets. The example images in Table \ref{tbl:pixel_dataset} mainly capture close-up views of the road surface taken at right angles, while the example images in Table \ref{tbl:pixel_dataset} showcasing bounding box datasets have a much wider field of view, including the vehicle's front. Therefore, the availability of bounding box datasets is crucial for developing algorithms that can detect cracks in a wider range of scenarios.

The RDD2018 dataset \cite{maeda2018road} is a collection of 9,053 road damage images captured using a smartphone mounted on a car. The dataset includes 15,435 instances of road surface damage that were captured under different weather and illuminance conditions. Each image in the dataset has a resolution of 600$\times$600, and is annotated with a bounding box that locates the damage on the road surface and identifies its type.

The RDD2019 dataset \cite{maeda2021generative} is an updated version of RDD2018 \cite{maeda2018road}. The dataset underwent a thorough review, relabelling and new annotated images were added to create RDD2019. This updated version contains a total of 13,135 images, which is an increase from 9,053 in RDD2018, and has 30,989 annotations, which is a substantial increase from 15,435 annotations in RDD2018. RDD2019 includes road images captured in Japan, and each image is annotated with a bounding box that locates the damage on the road and its type.

RDD2020 \cite{arya2020global} is a large-scale dataset that comprises 26,336 road images captured in India, Japan, and the Czech Republic. The dataset includes over 31,000 instances of road damage and is designed for developing deep learning-based methods to automatically detect and classify various types of road damage. RDD2020 covers a range of common road damage types, such as longitudinal cracks, transverse cracks, alligator cracks, and potholes. The images in RDD2020 were captured using cost-effective vehicle-mounted smartphones, enabling a  low-cost method for monitoring road pavement surface conditions. The dataset includes images with resolutions of either 600$\times$600 or 720$\times$720, providing a detailed view of the road surface. RDD2020 can be a useful resource for municipalities and road agencies seeking to develop cost-effective methods for monitoring and maintaining road conditions.

RDD2022 \cite{arya2022rdd2022} is the latest and most extensive version of the RDD datasets. Compared to its predecessors, RDD2022 offers the largest number of image samples, including 47,420 road surface images that contain 55,007 instances of road defects. Additionally, the dataset includes images with various resolutions, such as 512$\times$512, 600$\times$600, 712$\times$712, and 3650$\times$2044, which enables researchers to evaluate the performance of algorithms across different resolutions. RDD2022 provides an essential resource for researchers working on developing and training machine learning algorithms to detect and classify different types of road defects accurately. 

It is worth noting that several bounding-box-annotated datasets have been used for road defect detection studies; however, this survey paper focuses on providing information on publicly available datasets only.

\begin{table*}
\caption{Publicly available datasets for road surface defect detection.}
\label{tbl:bbox_dataset}
\centering
\setlength{\leftmargini}{0.4cm}
\resizebox{\textwidth}{!} 
{ 
\begin{tabular}{c|c|c|c|c|c|m{4cm}|c }
\toprule
\toprule
Name  & Format & Sensor  & Resolution & Label type & Surface & Description & Sample \\
\midrule      
\makecell{RDD2018\\ \cite{maeda2018road}}   &   RGB    &  \makecell{LG Nexus 5Xa}       &  \makecell{600$\times$600}
           &  \makecell{Bounding boxes\\(Multi-classes)}         &  \makecell{Asphalt}    & \begin{itemize}
       \item 9,053 road damage images
       \item 15,435 instances of road surface damage
   \end{itemize}   & \adjustbox{valign=c}{\includegraphics[width=20mm, height=20mm]{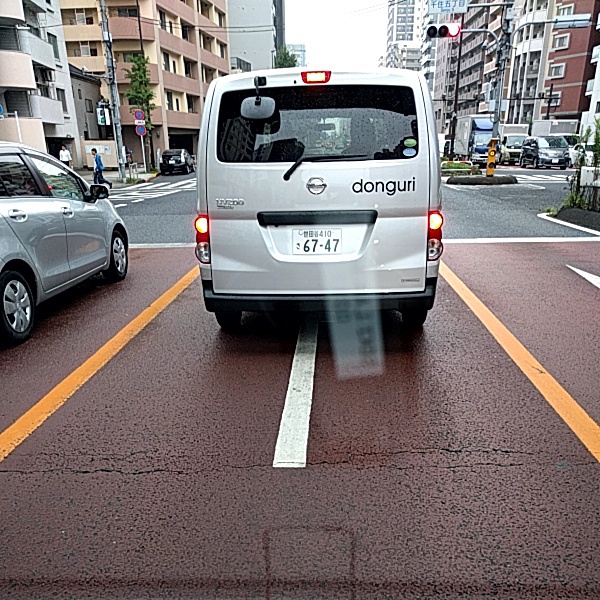}}  \\ \midrule
   
    \makecell{RDD2019\\ \cite{maeda2021generative}}   &   RGB    &  \makecell{In-vehicle\\camera}       &  \makecell{600$\times$600}
           &  \makecell{Bounding boxes\\(Multi-classes)}         &  \makecell{Asphalt}    & \begin{itemize}
       \item 13,135 road images 
       \item 30,989 instances of road damages
       \item Nine categories, including road damages and wheel marks
   \end{itemize}   & \adjustbox{valign=c}{\includegraphics[width=20mm, height=20mm]{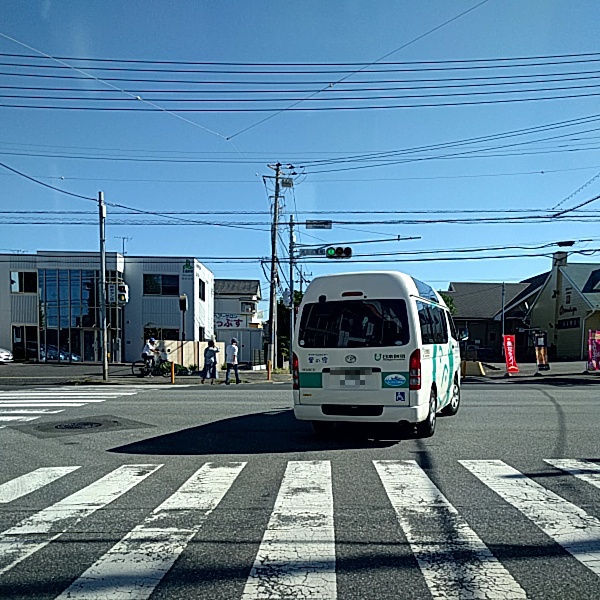}}  \\ \midrule

    \makecell{RDD2020\\\cite{arya2020global}}   &   RGB    &  \makecell{LG Nexus 5X\\Samsung Galaxy J6}       &  \makecell{600$\times$600\\720$\times$720}
           &  \makecell{Bounding boxes\\(Multi-classes)}         &  \makecell{Asphalt}    & \begin{itemize}
       \item 26,336 road images collected from India, Japan, and the Czech Republic
       \item 31,000 instances of road damages
       \item Four damage categories: Longitudinal Cracks, Transverse Cracks, Alligator
Cracks and Potholes.
   \end{itemize}   & \adjustbox{valign=c}{\includegraphics[width=20mm, height=20mm]{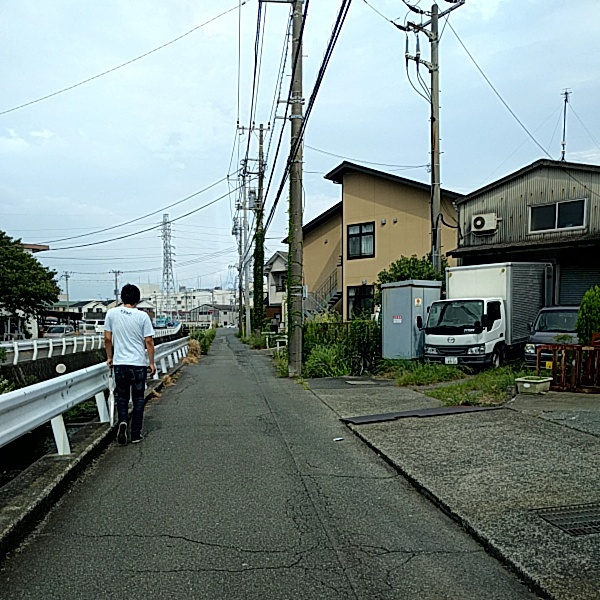}}  \\ \midrule

    \makecell{RDD2022\\\cite{arya2022rdd2022}}   &   RGB    &  \makecell{Smartphones\\High-resolution
Cameras\\Google Street View}       &  \makecell{512$\times$512\\600$\times$600\\720$\times$720\\3650$\times$2044}
           &  \makecell{Bounding boxes\\(Multi-classes)}         &  \makecell{Asphalt}    & \begin{itemize}
       \item 47,420 road images from Japan, India, the Czech Republic, Norway, the United States, and China 
       \item 55,000 instances of road damage
       \item Four damage categories: Longitudinal Cracks, Transverse Cracks, Alligator
Cracks and Potholes.
   \end{itemize}   & \adjustbox{valign=c}{\includegraphics[width=20mm, height=20mm]{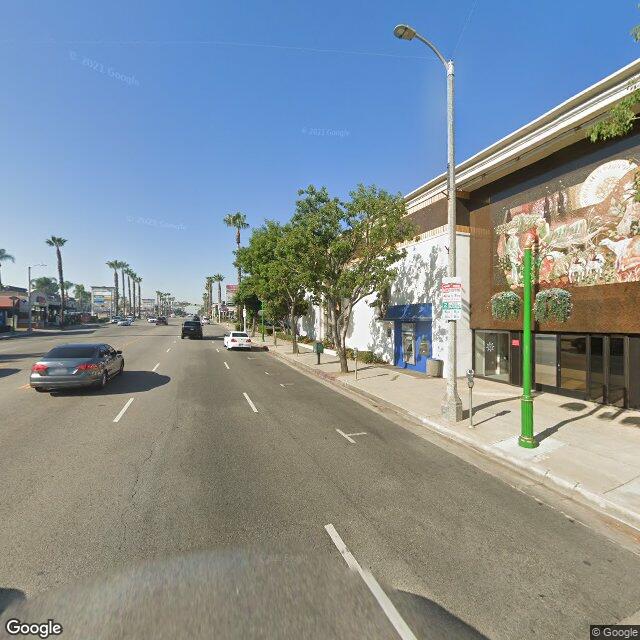}}  \\

\bottomrule
\bottomrule
\end{tabular}
}
\vspace{-2ex}
\end{table*}

\begin{table*}
\caption{Publicly available datasets containing depth information for road surface defect segmentation.}
\label{tbl:depth_dataset}
\centering
\setlength{\leftmargini}{0.4cm}
\resizebox{\textwidth}{!} 
{ 
\begin{tabular}{c|c|c|c|c|c|m{4cm}|c }
\toprule
\toprule
Name  & Format & Sensor  & Resolution & Label type & Surface & Description & Sample \\
\midrule          
\makecell{SHREC2022\\\cite{thompson2022shrec}}   &   RGB+D    &  \makecell{Luxonis OAK-D\\camera }       &   \makecell{Multiple\\resolutions}      &  \makecell{Pixel-wise\\(Binary class) }        &  \makecell{Asphalt}    & \begin{itemize}
       \item 4340 image pairs (RGB image and segmentation mask)
       \item 797 non-annotated RGB-D video clips
       \item Depth is estimated by disparity map
   \end{itemize}   & \adjustbox{valign=c}{\includegraphics[width=20mm, height=20mm]{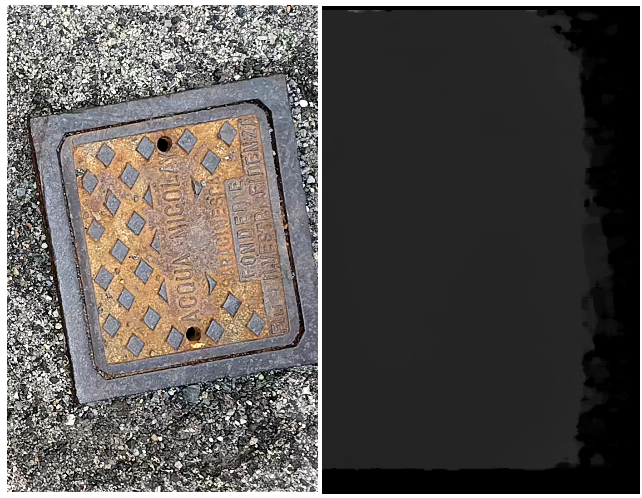}}  \\ \midrule
                  
        \makecell{Pothole-600\\ \cite{fan2019pothole}}   &   RGB+D    &  \makecell{ZED stereo\\camera }       &  400$\times$400     &  \makecell{Pixel-wise\\(Binary class) }        &  \makecell{Asphalt}    & \begin{itemize}
       \item 600 of image and depth map pairs
       \item Depth is estimated by disparity map
   \end{itemize}   & \adjustbox{valign=c}{\includegraphics[width=20mm, height=20mm]{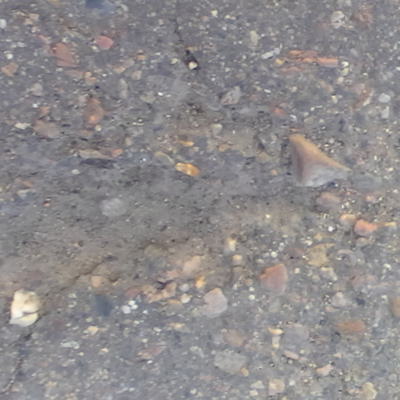}}  \\ 
   
\bottomrule
\bottomrule
\end{tabular}
}
\vspace{-2ex}
\end{table*}

\subsection{Depth sensor-based datasets}
Recently, many defect detection studies have utilised explicit depth sensors such as laser scanners and LiDAR, resulting in the release of various datasets containing point clouds or depth maps. Point clouds consist of a discrete set of data points in space and are usually represented by a list of 3D points. In contrast, a depth map can be considered as a structured point cloud, where the structure is more similar to that of an image frame, and each pixel value defines a depth value. Compared to point clouds, which are unstructured and only provide the coordinates of each point, depth maps offer a more structured representation of the data. Some point cloud datasets also contain colour codes for each point. These datasets provide valuable resources for researchers developing algorithms to process and analyse point cloud data for defect detection and other applications.

The SHREC 2022 dataset \cite{thompson2022shrec} is a comprehensive dataset created specifically for semantic segmentation of potholes and cracks on road surfaces. The dataset consists of 4,340 image/mask pairs, which were compiled from 5 publicly available datasets and augmented by a small number of manually segmented images. The images in the dataset vary in resolution and have been divided into training, validation, and test sets in 77/11/12 ratios. Furthermore, the dataset includes 797 non-annotated RGB-D video clips, each containing an RGB video and a disparity map video. Participants can extract additional images from these clips to enhance the dataset. 

The Pothole-600 dataset \cite{fan2019pothole} was collected using a ZED stereo camera. Three datasets containing 67 pairs of stereo images were created using Search Range Propagation (SRP) to estimate the road disparity images \cite{fan2018road}. The image resolutions of datasets 1, 2 and 3 are 1028 × 1730, 1030 × 1720, and 1028 × 1710 pixels, respectively, resulting in a total of 79 potholes captured. 

Overall, the above depth-sensor-based datasets are a valuable resource for researchers and practitioners working on the problem of semantic segmentation of road surface potholes and cracks. However, on the other side, to our best knowledge, there are currently no publicly available datasets that exclusively consist of point cloud data. Instead, a few datasets have been released that include both images and their corresponding depth maps. Although several methods using point clouds have been proposed, these datasets have not been publicly released. Unfortunately, there is a lack of publicly available datasets for point cloud analysis, which hinders the development and evaluation of new methods in this field.

%% file: 03_method.tex
\begin{figure*}
	\begin{center}
        % \fbox{\rule{0pt}{2in} \rule{\linewidth}{0pt}}
        \includegraphics[width=0.8\textwidth]{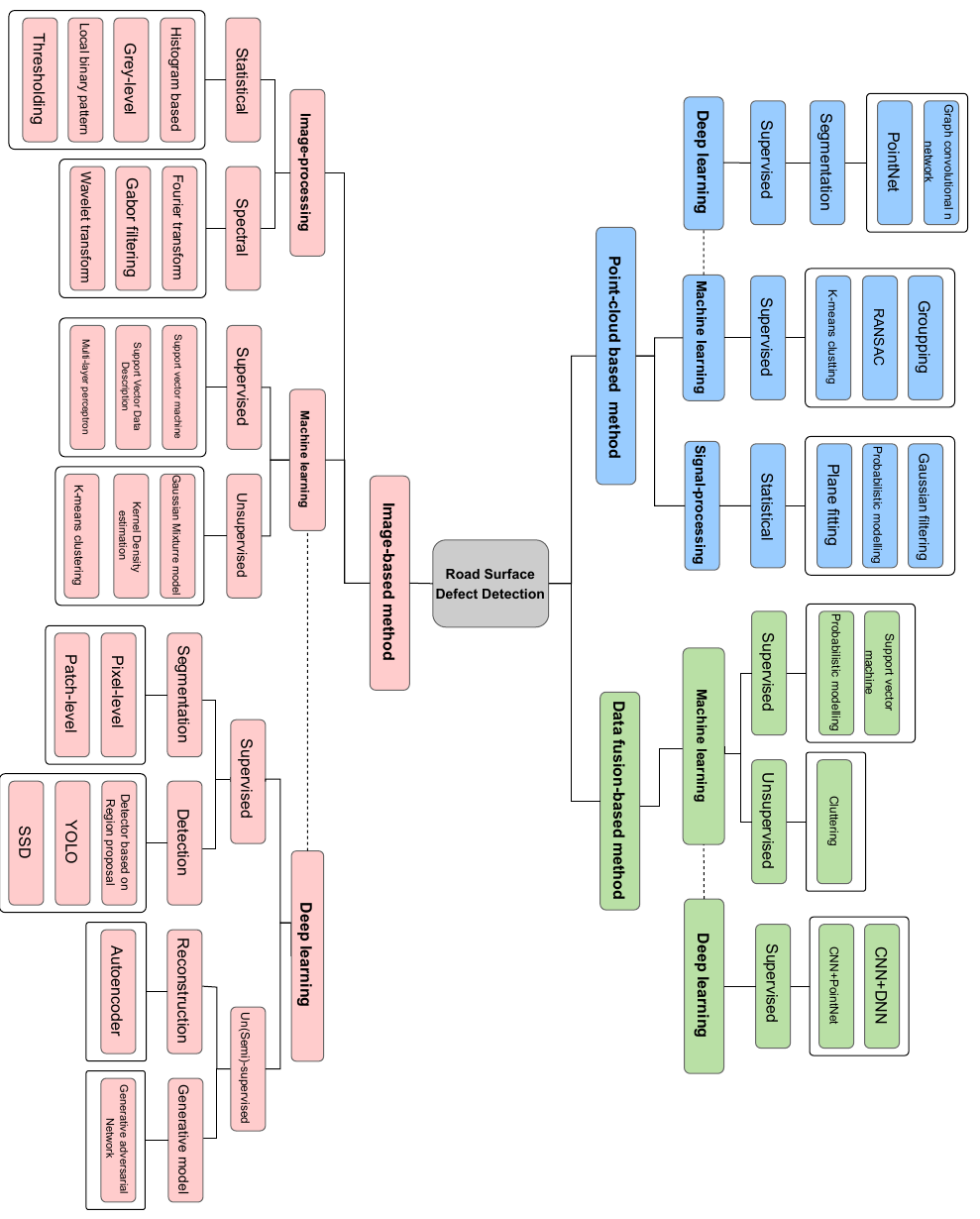}
           \caption{The taxonomy of road surface defect detection methods is structured as follows: The red-coloured categories represent the methodologies that use RGB images. The blue-coloured categories indicate the class of point cloud-based methods. The green-coloured categories define the methods using RGB images and depth values simultaneously. Deep learning is a sub-category of machine learning. However, since many methods based on deep learning have been proposed, it is connected to machine learning methods by dotted lines to represent methodological independence.}
           \end{center}
        \label{fig:taxonomy}
	\vspace{-4ex}
\end{figure*}

\section{Road surface defect detection}
\label{sec:5}
This section comprehensively reviews the various methods for detecting road surface defects. Due to the limited number of studies available on this topic, we refer to various reviews related to defect detection, such as texture fault detection \cite{timm2011non} and flat steel crack detection \cite{kumar2008computer}. Most researchers in the field mainly rely on image-based techniques that use RGB images to detect defects on material surfaces. They typically group defect detection techniques into different categories based on certain characteristics, although the exact taxonomy can vary depending on the researcher. One commonly used categorisation scheme is based on the type of approaches used. For example, Timm \etal \cite{timm2011non} broadly divides techniques for detecting texture faults into two categories: local and global. Meanwhile, defect detection techniques are often classified as classification-based, local-abnormality-based, and template-matching-based methods. Another categorisation scheme used by Youkachen \etal \cite{youkachen2019defect} groups defect detection techniques into statistical, network-based, proximity-based, deviation-based, and statistical models. 

The reviews mentioned above primarily focus on methodologies for defect detection using image-based methods, and their taxonomy was limited to this type of input data. However, in this paper, we aim to broaden the taxonomy by not only considering categories for classifying the defect detection methodologies but also considering the types of input data used. Specifically, we classify road surface defect detection methods based on their input data type, \ie image-based methods, point cloud-based methods, and data fusion-based methods. We then categorise these methods based on their underlying methodologies, including image processing-based, machine learning-based, and deep learning-based methods. To visualise this taxonomy structure, Fig. \ref{fig:taxonomy} provides an overview of the road surface defect detection classification scheme. By incorporating both the input data types and the underlying methodologies used, our taxonomy provides a more comprehensive framework for understanding and comparing different approaches to road surface defect detection.

\subsection{Image-based methods}
Compared to other sensors, such as laser scanners and LiDARs, cameras are a relatively inexpensive way to capture road surface images. As a result, RGB images are the most commonly used input data for road surface defect detection. We categorise image-based methods into three groups: image-processing-based, machine-learning-based, and deep-learning-based. In this section, we provide detailed descriptions of each of these methods. Image-processing-based methods typically involve using various algorithms to extract features from road surface images, which are then used to detect defects. Machine learning-based methods involve training a model on a dataset of labelled road surface images to learn patterns associated with different types of defects. Deep learning-based methods use neural networks with multiple layers to automatically learn features and patterns from raw images. Table \ref{tbl:image-based-methods} provides an overview of the most commonly used image-based methods for road defect detection.

\subsubsection{Image processing-based approaches}
The group of image-processing-based methods encompasses a variety of techniques that leverage statistical or spectral methods to analyse digital images. In this section, we introduce four statistical methods and three spectral methods, providing an overview of each.

\noindent
\textbf{Statistical:} Statistical methods for surface defect detection can be broadly classified into four categories: thresholding-based, local binary pattern-based, grey-level-based, and histogram-based methods. Among these, thresholding methods are the most commonly used for detecting defective regions on flat surfaces and are widely employed in Automated Visual Inspection (AVI) systems. However, traditional thresholding methods that rely on a fixed pixel value for defect detection are susceptible to random noise and non-uniform illuminations. To overcome this limitation, Djukic and Spuzic \cite{djukic2007statistical} proposed an adaptive thresholding approach that estimates the probability distribution of pixel intensities. This dynamic thresholding approach accurately separates true defects from random noise, and has been successfully applied in several real-world scenarios.

The local binary pattern (LBP) operator is another widely used classical technique for characterising local texture features of images. One of the main advantages of LBP is its invariance to rotation and grey-level changes. Initially proposed in 1994 by Ojala \etal \cite{ojala1996comparative}, LBP has been frequently used for detecting defects on flat steel surfaces \cite{chu2017steel,liu2018online,zhao2018steel}. However, the original LBP has some limitations, including weak global descriptive ability and sensitivity to noise. To address these issues, several LBP variants have been developed by modifying the threshold or scale of the original LBP. These variants have been widely used in defect detection of flat steel surfaces. For example, Wang \etal \cite{wang2019surface} proposed an LBP-inspired feature extractor that estimates the variations in four directions, including horizontal, vertical, and two diagonal directions, simultaneously. LBP variants can be applied as lightweight feature descriptors in both defect detection and classification tasks. Looking ahead, future trends in automated visual inspection systems involve the development of more noise-robust and scale-invariant LBP variants or LBP-like descriptors. This would enable more accurate and efficient detection of defects in a wide range of materials and conditions.

Grey-level statistics is an alternative approach that can be used for defect detection. Direct thresholding techniques may prove ineffective for defect detection in low-contrast images, making it necessary to analyse the distribution of grey levels in the image before performing the thresholding operation. Yang \etal \cite{yang2008method} utilised features such as the mean value and distribution of pixels from the steel surface background to simultaneously differentiate between bright and dark defect objects. To ensure insensitivity to noise, Choi and Kim \cite{choi2012unsupervised} employed a spectral-based approach to estimate the distribution of the background and then refined the defective regions locally to obtain probabilistic estimates. This method is superior to previous defect detection methods and provides robust outcomes even in noisy environments. However, the applicability of these methods for surface defect detection is limited by the diversity of surface defects encountered in various scenarios. Therefore, developing more sophisticated grey-level statistics-based methods is necessary to effectively handle various types of surface defects encountered in different scenarios. These methods can potentially offer more accurate and reliable results in defect detection and classification tasks, especially in low-contrast images.

\noindent
\textbf{Spectral:} Despite statistical approaches that have been extensively researched for surface defect detection, they may not be reliable for detecting subtle defects with low-contrast intensity transitions. This is particularly true when illumination changes or pseudo defects are frequently present. As a result, there is a growing need for emerging Automated Visual Inspection (AVI) methods to detect surface defects in real-world environments. 

Spectral approaches can be broadly classified into transform-based and filter-based methods. Transform-based methods employ Fourier Transform (FT) to achieve translation, expansion, and rotation invariance of image features. However, images from the real world often require additional processing to enhance image quality, such as denoising and contrast enhancement. For instance, longitudinal cracks on complex backgrounds can be detected by computing features with translational invariance using the Fourier amplitude spectrum of each subband \cite{ai2013surface}. Another unsupervised method based on Phase-Only Transform (PHOT) was proposed in \cite{aiger2010phase} inspired by discrete FT, effectively highlighting irregular patterns to indicate defects. This novel method demonstrated its effectiveness and versatility on various textured surfaces such as wood, steel, ceramic, and silicon wafers. Although transform-based methods have numerous advantages in terms of translation, expansion, and rotation invariance, they may have limitations in handling noise and illumination variation.

Wavelet transform offers the advantage of adjusting the time-frequency window and adapting it with the frequency change in the window's centre. Additionally, the characteristics of wavelets are more compatible with the human visual system, enabling the extraction of information from signals and the execution of multiscale analysis of functions or signals through scaling and shifting operations. Therefore, wavelet-based methods have been extensively used for image analysis, including surface defect detection.

Filter-based methods operate in the frequency domain to represent an image, which can lead to the neglect of local descriptive information in the spatial domain. This limitation can be addressed by Gabor filters, which modulate a Gaussian kernel function on a sinusoidal wave with a specific frequency, resulting in both spatial and frequency domain analysis \cite{xie2010fusing}. A simple 2-D Gabor filter can achieve localised and oriented frequency analysis \cite{kim2017fast}. However, it is important to note that Gabor filters can be computationally expensive, and there may be better options available for optimisation, such as Finite Impulse Response (FIR) filters, which offer more freely adjustable parameters compared to Infinite Impulse Response (IIR) and Gabor filters, as suggested in \cite{kumar2001automated}.

\begin{table*}
\caption{List of image-based methods.}
\centering
\resizebox{\textwidth}{!} 
{ 
	\begin{tabular}{c|c|c|c|c|c|m{8cm}}
		\toprule
		Cat & Subcat & Reference & Year  & Based method  & Output protocol  & \multicolumn{1}{c}{Details} \\
		\hline
		
		\multirow{6}{*}{\rotatebox[origin=c]{90}{Image processing-based method}} & \multirow{3}{*}[-6ex]{\rotatebox[origin=c]{90}{Statistical}}  & Djukic and Spuzic \cite{djukic2007statistical} & 2007 & Kernel estimation  & \makecell{Pixel-level\\(Binary)}  &  
		Estimate the probability distribution of pixel intensities and apply adaptive thresholding to detect defects.
		\\ \cline{3-7}
  
        &  & Wang \etal \cite{wang2019surface} & 2019 &  LBP  & \makecell{Patch-level\\(Binary)}  &  
		Propose lightweight feature descriptor. Estimating the variations of four directions simultaneously, which are horizontal, vertical, and two diagonal directions, so that the features are more robust to directional variation.
		\\ \cline{3-7}

        &  & Yang \etal \cite{yang2008method} & 2008 & Gray-level statistic   & \makecell{Pixel-level\\(Binary)}  &  
		Utilise the statistics of features \ie mean value and distribution of pixels. Separate bright and dark areas on a road surface to distinguish between normal road surfaces and defects.
		\\ \cline{2-7}

    & \multirow{2}{*}{\rotatebox[origin=c]{90}{Spectral}} & Aiger and Talbot \cite{aiger2010phase} & 2010 & Fourier transform   & \makecell{Patch-level\\(Binary)} &  
		Propose an unsupervised method based on the Fourier transform. The proposed method can persist only irregular patterns to present defects.
		\\ \cline{3-7}

   &   & Liu and Yan \cite{liu2014automated}  & 2014 & Wavlet transform  & \makecell{Patch-level\\(Binary)} &  
		Present a novel wavelet-based image filtering algorithm based on anisotropic diffusion for detecting road defect detection.
		\\ \cline{1-7}

		\multirow{6}{*}[-4ex]{\rotatebox[origin=c]{90}{Machine learning-based method}} & \multirow{3}{*}[-2ex]{\rotatebox[origin=c]{90}{Supervised}}  & Lin and Liu \cite{lin2010potholes} & 2010 & SVM  & \makecell{Patch-level\\(Binary)}  &  
		Distinguish whether a cropped image patch contains a pothole or a normal road surface. 
		\\ \cline{3-7}
  
        &  & Saar \etal \cite{saar2010automatic} & 2010 &  Multi-layer Perceptron  & \makecell{Patch-level\\(Binary)}  &  
		Derive neural networks which can provide patch-level defect detection. The neural network is trained based on hand-crafted features.
		\\ \cline{3-7}

        &  & Yusof \etal \cite{yusof2019automated} & 2019 & Multi-layer perceptron   & \makecell{Patch-level\\(Binary)} &  
		Similar to Saar \etal \cite{saar2010automatic}, deploy the defect detection method based on multi-layer perceptron. Hand-crafted features are used to train parameters.
		\\ \cline{2-7}

    & \multirow{3}{*}[-3ex]{\rotatebox[origin=c]{90}{Unsupervised}} & Oliveira \etal \cite{oliveira2012automatic} & 2012 & Gaussian Mixture Model   & \makecell{Patch-level\\(Binary)}  &  
		Model the probabilistic distribution of features extracted from normal road surfaces using multiple Gaussian distributions. Detect defects using a pre-defined threshold. 
		\\ \cline{3-7}

   &   & Buza \etal \cite{buza2013pothole}  & 2013 & Clustering  & \makecell{Patch-level\\(Binary)}  &  
		Leverage K-means clustering to compile a normal road surface model. Detect road defects by computing the distance between the input feature and centroids.
		\\ \cline{3-7}

   &  & Oliveira \etal \cite{oliveira2012automatic} & 2012 & Kernel density estimation   & \makecell{Pixel-level\\(Binary)}  &  
		Directly estimate a probabilistic model for the normal road surface using kernel density estimation. Detect road defects using a pre-defined threshold.
		\\ \cline{1-7}

		\multirow{10}{*}[-14ex]{\rotatebox[origin=c]{90}{Deep learning-based method}} & \multirow{5}{*}[-20ex]{\rotatebox[origin=c]{90}{Supervised}}  & Zou \etal \cite{zou2018deepcrack} & 2018 & U-Net  & \makecell{Pixel-level\\(Binary)}  &  
		Pixel-level road defect segmentation model. Provide binary classification results representing likelihoods for normal road surfaces and defects. 
		\\ \cline{3-7}
  
        &  & Konig \etal \cite{konig2019convolutional} & 2019 &  U-Net  & \makecell{Pixel-level\\(Binary)}  &  
		Utilise an attention mechanism and residual convolutional blocks to improve discriminative power for the defect segmentation model.
		\\ \cline{3-7}

        &  & Zhang \etal \cite{zhang2020crackgan} & 2020 & GAN   & \makecell{Patch-level\\(Binary)}  &  
		Propose a deep generative adversarial network for pavement crack detection, which is trained end-to-end with a crack-patch-only supervised method to overcome the local minimum problem. 
		\\ \cline{3-7}

   &  & Chen \etal \cite{chen2017nb} & 2017 & CNN   & \makecell{Pixel-level\\(Binary)} &  
		Present a CNN architecture for pixel-level defect detection based on video data. a Naive Bayes data fusion scheme is proposed to reduce false positives.
		\\ \cline{3-7}

    &  & Dong \etal \cite{dong2022automatic} & 2023 & CNN   & \makecell{Pixel-level\\(Binary)} &  
		propose a pavement damage segmentation framework consisting of a Siamese network,  an explicit crack segmentation network, and a feature fusion method.
		\\ \cline{3-7}

   &  & Chen \etal \cite{chen2023multi} & 2023 & CNN   & \makecell{Pixel-level\\(Binary)} &  
		Present a CNN architecture using an attention mechanism for multi-scale defect detection. The multi-scale attention method improves the robustness of defect detection methods for various image scales.
		\\ \cline{3-7}

  &  & Wang \etal \cite{wang2023normal} & 2023 & CNN   & \makecell{Pixel-level\\(Binary)} &  
		propose a relevance-aware and cross-reasoning network (RCN) for road defect segmentation, which can segment defects using merely non-defective images for training.
		\\ \cline{3-7}

   &  & Fang \etal \cite{fang2020novel} & 2020 & RCNN   & \makecell{Instant-level\\(Multi-classes)}  &  
		Present instant-level road defect detection method. The proposed method is based on Faster-RCNN (Region-based Convolutional Neural Networks). Provide detection results identified by multiple defect classes. 
		\\ \cline{2-7}

    & \multirow{3}{*}[-4ex]{\rotatebox[origin=c]{90}{Unsupervised}} & Mujeeb \etal \cite{mujeeb2019one} & 2019 & Autoencoder   & \makecell{Pixel-level\\(Binary)}  &  
		Train an autoencoder using normal road surface images. Detect road defects by comparing a reconstruction error and a pre-defined threshold.
		\\ \cline{3-7}

  & & Yu \etal \cite{yu2020unsupervised} & 2020 & GAN  & \makecell{Pixel-level\\(Binary)}  &  
		Build an adversarial network to derive a transformation model between a normal road surface and its corresponding frequency-domain image. Defects are detected by comparing the difference in the frequency domain. 
		\\ \cline{3-7}

   &   & Yu \etal \cite{yu2023multi} & 2023 & U-Net  & \makecell{Pixel-level\\(Binary)}  &  
		Design unsupervised defect detection using based on multi-source domain adaptation. Provide more stable and improved segmentation performance on an unlabelled dataset. 
		\\ \cline{1-7}

      & \multirow{3}{*}[-4ex]{\rotatebox[origin=c]{90}{Semi-supervised}} & Li \etal \cite{li2020semi} & 2020 & GAN   & \makecell{Pixel-level\\(Binary)}  &  
		A semi-supervised method for pavement crack detection. An adversarial learning method and a full convolution discriminator are adopted, which can learn to distinguish the ground truth from segmentation predictions.
		\\ \cline{3-7}

  & & Shim \etal \cite{shim2020multiscale} & 2020 & GAN  & \makecell{Pixel-level\\(Binary)}  &  
		Multiscale and adversarial learning-based semi-supervised semantic segmentation approach for crack detection in concrete surface. The multiscale segmentation neural network, discriminator neural network, and adversarial learning technique were used.
		\\ \cline{3-7}

   &   & Ren \etal \cite{ren2023semi} & 2023 & GAN  & \makecell{Pixel-level\\(Binary)}  &  
		A pavement anomaly detection network (PAD Net) is proposed. PAD Net is a semi-supervised learning approach based on generative adversarial networks for identifying pixel-level anomalous image segments.
		\\ \cline{1-7}
		\bottomrule
	\end{tabular}
	}
	\label{tbl:image-based-methods}
%\vspace{-1em}
\end{table*}

\subsubsection{Machine learning-based methods}
The machine learning group comprises a range of approaches that utilise machine learning algorithms such as Support Vector Machine (SVM) \cite{lin2010potholes} or Neural Networks \cite{saar2010automatic,yusof2019automated} to analyse images based on hand-crafted features (in contrast to the features automatically learned by neural networks in deep learning based methods discussed later in Section~\ref{sec:DeepLearningMethods}). These approaches can be broadly categorised into two groups: supervised and unsupervised methods, depending on whether annotated data is available or not.

\noindent
\textbf{Supervised:} Supervised methods use explicit annotation to define whether the input image contains defect patterns or not. SVMs have been used in \cite{lin2010potholes} to distinguish between potholes or normal road surfaces in cropped image patches. Similarly, neural networks were used for path-level defect detection in \cite{saar2010automatic} and \cite{yusof2019automated}. In these methods, the input image is first divided into smaller patches, and hand-crafted features are extracted from each patch. The features are then used to train machine learning-based models, which are subsequently deployed to identify defects in the road surface. While these approaches demonstrated better performances than simple image-processing-based methods, the hand-crafted features they rely on may not be robust to changes in illumination conditions caused by weather or day-and-night variations. Moreover, these methods are limited in covering complex feature distributions.
 
\noindent
\textbf{Unsupervised:} Unsupervised methods aim to detect road defects without explicit annotation to represent them. Instead, these methods derive a model from representing normal road surface features \cite{oliveira2012automatic,abdel2006pca,oliveira2010improved,oliveira2009automatic,buza2013pothole}. They assume that the training data consists entirely of normal road surfaces, and based on these assumptions, the model is trained. If the model's output is abnormal, it is determined that there is a defect. These approaches' quality highly depends on how sophisticated the probability model can be derived. Kernel density estimation has been used to derive a probabilistic model for features extracted from normal road surface images \cite{oliveira2010improved,oliveira2009automatic}. Alternatively, Oliveira \etal \cite{oliveira2012automatic} propose an unsupervised road defect detection method based on the Gaussian mixture model. Buza \etal \cite{buza2013pothole} leverage k-means clustering to derive a normal road surface model. While these methods can be useful in detecting defects without the need for explicit annotation, they may suffer from the overfitting problem when dealing with complex and diverse road surface images. Furthermore, these approaches have an inherent weakness as they assume that the feature distribution follows a Gaussian or linear distribution, which may not always hold true for actual data. This can lead to limited effectiveness and accuracy of the models when the actual data distribution significantly deviates from the assumed distribution. Therefore, it is important to consider the choice of probability models carefully and to validate their performance on real-world datasets with diverse feature distributions.

%Because those unsupervised methods do not need explicit annotation for road defects, they can be thought of as a significant advantage for real-world situations. However, these approaches still have an inherent weakness which are their approaches work effectively under the assumption that the feature distribution has a Gaussian distribution or a linear distribution. However, this assumption is easily broken in the distribution of actual data.

\subsubsection{Deep learning-based methods}
\label{sec:DeepLearningMethods}
In this subsection, we will introduce deep learning-based road condition assessment from two perspectives: bounding box level detection and pixel level detection. It is worth noting that deep learning is a subset of machine learning that uses neural networks with multiple layers to learn representations of data. As such, deep learning methods are becoming increasingly popular in this field due to their ability to automatically learn features from data, reducing the need for hand-crafted feature engineering.

\noindent
\textbf{Supervised:} Supervised methods using deep learning can be classified into two categories: pixel-level defect detection and objective-level detection. Pixel-level defect detection involves defect segmentation, where each pixel in the image is classified as a defect or non-defect. The encoder-decoder network is one of the most famous structures used for defect segmentation. For example, in~\cite{jenkins2018deep}, the U-Net~\cite{ronneberger2015u}, a classical encoder-decoder network, was used to segment cracks on the road. 

Several recent studies have proposed novel deep-learning approaches for crack detection on the road surface. Zou \etal \cite{zou2018deepcrack} proposed a novel encoder-decoder network that separates cracks from the background using convolutional feature learning. Similarly, Pauly \etal \cite{pauly2017deeper} leverage convolutional neural network to obtain pixel-wise classification results between normal road surfaces and road defects. Konig \etal \cite{konig2019convolutional} introduced a U-Net-based network with an attention mechanism and residual convolutional blocks that achieved state-of-the-art results in two distinct crack datasets. Ji \etal \cite{ji2020integrated} utilised the DeepLabv3+ model \cite{chen2018encoder} for automatic crack detection in asphalt pavement. In addition, Zhang \etal \cite{zhang2020crackgan} proposed a deep generative adversarial network for pavement crack detection, which was trained end-to-end with a crack-patch-only supervised method to overcome the local minimum problem.  Dong \etal \cite{dong2022automatic} proposed a pavement damage segmentation framework consisting of a Siamese network and an explicit crack segmentation network. This framework has been extended by applying the feature fusion method \cite{dong2022innovative}. Wang \etal \cite{wang2023normal} proposed a Relevance-aware and Cross-Reasoning network (RCN) for road defect segmentation, which can segment defects using merely non-defective images for training. The aforementioned methods do not consider the robustness of the image scale. To improve the robustness of image scale variation, Chen \etal \cite{chen2023multi} leveraged the multi-scale attention method, which can detect road defect performance in various image resolutions.

Defect detection at the bounding box level can be achieved through either patch scanning or end-to-end Convolutional Neural Network (CNN) approaches. The patch scanning-based method involves cropping the original high-resolution image into a set of low-resolution patches, identifying patches containing defects using a trained classifier, and applying filtering to remove false positives. In \cite{chen2017nb}, a CNN architecture was employed to classify crack patches, and a registration procedure was proposed to maintain the spatiotemporal coherence of cracks in videos. A Naive Bayes data fusion scheme, which aggregates information from multiple frames, was proposed to effectively discard false positives. However, this approach requires a high density of patches, leading to increased computational load and challenging real-time applications. 

Some other works use CNN detectors to predict the bounding box of cracks directly \cite{ma2022automatic,ren2015faster,liu2019deepcrack,fang2020novel}. Whether a Region Proposal Network (RPN) is used, these detectors can be divided into two families: one-stage and two-stage. A typical two-stage network is Faster-RCNN (Region-based Convolutional Neural Networks)~\cite{ren2015faster}, which uses an RPN to generate regions of interest in the first stage, and then performs object classification and bounding-box regression on the region proposals. Li \etal \cite{liu2019deepcrack} used Faster R-CNN to detect six types of road defects under different lighting conditions.
Similarly, Fang \etal \cite{fang2020novel} employed Faster-RCNN to identify the candidate crack regions, followed by a Bayesian integration algorithm to filter out false positives. Ma \etal \cite{ma2022automatic} proposed a system composed of a Pavement Crack Generative Adversarial Network (PCGAN) and a crack detection and tracking network called YOLO-MF. In this method, First, The PCGAN is leveraged to generate virtual crack images. Next, YOLO-MF is developed to identify the regions of cracks.

Although two-stage networks can reduce some computational load, they are still not fast enough for real-time applications. Therefore, single-stage networks such as  You Only Look Once (YOLO)~\cite{redmon2016you} and Single Shot MultiBox Detector (SSD)~\cite{liu2016ssd}  have been used to improve perception speed significantly. Single-stage networks treat object detection as a simple regression problem by taking an input image and learning the class probabilities and bounding box coordinates.
% add several references
YOLOv2~\cite{redmon2017yolo9000} and YOLOv3~\cite{redmon2018yolov3} were used for crack detection of road surfaces in \cite{mandal2018automated} and \cite{park2020concrete}, respectively. These studies showed that cracks on structural surfaces can be detected and quantified with high accuracy in real-time when one-stage detectors are used.

\noindent
\textbf{Unsupervised and semi-supervised:} Several unsupervised methods based on autoencoders \cite{vincent2010stacked} have been proposed for road defect detection~\cite{mujeeb2019one, kang2018deep}. These methods train models on normal road images and detect defects by checking whether the reconstruction error of an input sample exceeds a threshold. However, this reconstruction-based approach has a significant drawback: the model may generate plausible-looking samples that were not seen during training, making it difficult to distinguish whether a sample contains defects or not \cite{perera2019ocgan,pidhorskyi2018generative}. To address this limitation, Yu \etal \cite{yu2020unsupervised} proposed an adversarial network-based road defect detection method for road defect detection. Their model uses a generator network to synthesise defect samples fed to a discriminator network that distinguishes between real and synthesised defect samples. The generator is trained to produce defect samples that can fool the discriminator, while the discriminator is trained to improve its ability to distinguish between real and synthesised samples. This adversarial training process leads to a model that can detect road defects even without labelled training data. As similar to the above method, Yu \etal \cite{yu2023multi} presented an unsupervised defect segmentation method based on multi-source domain adaptation. Their method improves the segmentation performance on an unlabelled dataset using multiple labelled datasets.

Unsupervised methods offer the advantage of not relying on labelled data, which makes them appealing to real-world environments. However, their actual performance is noticeably inferior to that of supervised methods \cite{yu2020unsupervised,pidhorskyi2018generative,vincent2010stacked}. Specifically, methods based on the fine-tuning technique, where a model is retrained using generated pseudo labels, exhibit varying performance depending on the accuracy of these labels \cite{yu2020unsupervised,yu2023multi}. To address these limitations, semi-supervised methods have been proposed \cite{li2020semi,shim2020multiscale,ren2023semi}.

Li \etal~\cite{li2020semi} and Shim \etal~\cite{shim2020multiscale} have applied adversarial learning in semi-supervised road defect detection. They employed adversarial learning to minimise domain shifts between a large-scale labelled dataset and a dataset with only a few labelled samples. This approach helps bridge the gap between labelled and unlabelled data to improve model performance. Another study by Ren \etal~\cite{ren2023semi} utilises adversarial learning to derive a normal road surface model, enhancing the accuracy of road defect detection. By leveraging semi-supervised techniques and adversarial learning, these methods aim to improve the performance and robustness of road defect detection, especially when labelled data is limited or unavailable. 

\textbf{Remarks.}
Image-based methods excel in road pavement defect detection, offering numerous advantages. They are non-invasive, ensuring the safety of inspection personnel by eliminating physical contact with the pavement. This results in efficient, large-scale coverage, surpassing manual inspection speeds. The objective nature of image-based systems ensures consistent results, free from human error and subjective judgment. Additionally, these methods facilitate collecting and analysing historical data, which is invaluable in understanding pavement deterioration patterns and aiding future planning.

However, these methods have notable drawbacks: Defect detection accuracy depends heavily on image quality and is susceptible to poor lighting, shadows, and adverse weather conditions; They primarily detect surface defects, missing critical subsurface issues; Furthermore, privacy concerns arise with imaging technology, especially if images are captured beyond the road surface. These challenges emphasise the need for careful implementation of such systems.

In summary, the past decades have witnessed the rapid development of visual road defect detection, with traditional image processing methods and deep learning methods being the two main approaches. Deep learning-based methods, in particular, have gained more attention due to their better robustness and accuracy. Despite the progress, there are still limitations in image-based road defect detection, such as the susceptibility of visual methods to varying lighting conditions and the high computational load required to process high-resolution images. Therefore, there is a need to address these challenges and improve the performance of image-based road defect detection. Two possible directions for future research are (1) enhancing the real-time performance of the algorithms and (2) designing an invariant feature extractor that can adapt to different environments. By addressing these challenges, image-based road defect detection can be further improved and applied in real-world scenarios.

\begin{table*}
\caption{List of point cloud-based methods.}
\centering
\resizebox{\textwidth}{!} 
{ 
	\begin{tabular}{c|c|c|c|c|c|m{8cm}}
		\toprule
		Cat & Subcat & Reference & Year  & Based method  & Output protocol  & \multicolumn{1}{c}{Details} \\
		\hline
		
		\multirow{4}{*}[-2ex]{\rotatebox[origin=c]{90}{Signal processing}} & \multirow{4}{*}[-10ex]{\rotatebox[origin=c]{90}{Statistical}}  & Xu \etal \cite{xu2019intelligent} & 2019 & Gaussian filtering  & \makecell{Point-level\\(Binary)}  &  
		Identify cracks from the point clouds measurement and optimise the accuracy of the crack analysis efficiently using Gaussian filtering
		\\ \cline{3-7}
  
        &  & Yu \etal \cite{yu20143d} & 2014 &  Probabilistic modelling  & \makecell{Point-level\\(Binary)}  &  
		Modelling the intensity information of point clouds and using the intensity model to identify pavement cracks and extract their skeletons. 
		\\ \cline{3-7}

        &  & Li \etal \cite{li20173d} & 2017 & Point clustering   & \makecell{Point-level\\(Binary)}  &  
		 laser-imaging techniques are employed to model the pavement surface with dense 3D points, and a sparse points grouping method is proposed to detect cracks from the 3D point clouds. 
		\\ \cline{3-7}

        &  & Bosurgi \etal \cite{bosurgi2023automatic} & 2023 & Thresholding   & \makecell{Point-level\\(Binary)}  &  
		 propose an approach for automated pothole detection through the processing of depth sensing results of pavement surfaces acquired using innovative high-performance equipment.
		\\ \cline{3-7}

   &  & Du \etal \cite{du2020pothole} & 2020 & Plane fitting   & \makecell{Point-level\\(Binary)}  &  
		3D point clouds are first acquired by using binocular stereo vision, and then a pavement plane fitting method is used to eliminate the 3D point clouds of the road plane and roughly extract the pothole.
		\\ \cline{1-7}

		\multirow{3}{*}[-2ex]{\rotatebox[origin=c]{90}{Machine learning}} & \multirow{3}{*}[-4ex]{\rotatebox[origin=c]{90}{Supervised}}  & Wu \etal \cite{wu2016rapid} & 2016 & Clustering  & \makecell{Point-level\\(Binary)}  &  
		Extract 3D crack skeletons from 3D point clouds acquired by a mobile Light Detection and Ranging (LiDAR) system. The algorithm leverages the clustering method to distinguish whether a point is a crack or a normal road surface.
		\\ \cline{3-7}
  
        &  & Yan \etal \cite{yan2017detection} & 2017 &  RANSAC  & \makecell{Point-level\\(Binary)}  &  
		The proposed method assumes that the majority of scanned points from a road surface can be modelled well using RANSAC, and points extracted from road defects can be considered outliers by using the RANSAC algorithm. 
		\\ \cline{3-7}

        &  & Yu \etal \cite{yu20143d} & 2014 & K-means clustering   & \makecell{Point-level\\(Binary)} &  
		Conduct binary clustering to identify 3D points extracted from normal road surfaces and defects. The clustering is conducted using depth value.
		\\ \cline{1-7}

		\multirow{7}{*}[-10ex]{\rotatebox[origin=c]{90}{Deep learning}} & \multirow{6}{*}[-20ex]{\rotatebox[origin=c]{90}{Supervised}}  & Dizaji \etal \cite{dizaji20193d} & 2019 & 3D InspectionNet  & \makecell{Voxel-level\\(Binary)}  &  
		Learn 3D defect features from an artificially generated 3D dataset. Distinguish defect features extracted from the voxels. Voxels are defined from point clouds.
		\\ \cline{3-7}
  
        &  & Zhang \etal \cite{zhang2017automated} & 2017 &  CrackNet  & \makecell{Point-level\\(Binary)}  &  
		A deep learning-based method that detects cracks based on 3D asphalt surface data. Provide point-wise 
 binary class scores for all individual points. This job is extended to CrackNet II \cite{zhang2018deep} and  CrackNet-V \cite{fei2019pixel}.
		\\ \cline{3-7}

        &  & Nasrollahi \etal \cite{nasrollahi2019concrete} & 2019 & PointNet   & \makecell{Point-level\\(Binary)}  &  
		Directly use point clouds as input without pre-processing, such as voxelisation or making a histogram. The proposed method provides defect detection on a concrete surface.
		\\ \cline{3-7}

  &  & Feng \etal \cite{feng2021gcn} & 2021 & Graph Convolutional   & \makecell{Point-level\\(Binary)}  &  
		A graph-widen module has been proposed to construct a reasonable graph structure for point clouds on graph convolutional networks (GCN) for increasing the detection performance of road defects. 
		\\ \cline{3-7}

  &  & Li \etal \cite{li20223d} & 2023 & Mask R-CNN   & \makecell{Point-level\\(Binary)}  &  
		A pipeline for accurately segmenting and extracting rural road surface objects in 3D LiDAR point-cloud is proposed. As well as a method to extract geometric parameters belonging to tar seal. 
		\\ \cline{3-7}

   &  & Li \etal \cite{li2021classification} & 2021 & PointNet++   & \makecell{Point-level\\(Multi-classes)} &  
		Propose PointNet++-base road surface defect classification method. The proposed method takes raw point-cloud without modification. The proposed method adaptively combines features from multiple scales. 
		\\ \cline{3-7}

  &  & Chen \etal \cite{chen2023rapid} & 2023 & GCN, PointNet++   & \makecell{Point-level\\(Binary-class)} &  
		A deep learning-based gradation extraction method using 3D point cloud data is proposed. A multi-feature fusion network was developed, which uses an extraction network as the backbone and additional auxiliary information.
		\\ \cline{3-7}

  &  & Gui \etal \cite{gui2023transfer} & 2023 & Deep Neural Network   & \makecell{Point-level\\(binary-classes)} &  
		Inspired by the heterogeneity of rapid-increasing 3D pavement data and the generalisation ability of transfer learning, a robust and generalised framework for cross-scene 3D pavement-crack detection and attribute extraction was proposed.
		\\ \cline{2-7}

    & \multirow{1}{*}{Semi-supervised}  & Feng \etal \cite{feng2021gcn} & 2021 & Graph Convolutional   & \makecell{Point-level\\(Binary)}  &  
		A graph-widen module has been proposed to construct a reasonable graph structure for point clouds on graph convolutional networks (GCN) for increasing the detection performance of road defects. 
		\\ \cline{1-7}
		\bottomrule
	\end{tabular}
	}
	\label{tbl:point-based-methods}
%\vspace{-1em}
\end{table*}

\subsection{Point cloud-based methods}
In recent years, various approaches have been proposed for depth-based methods, such as point clouds or depth maps, utilising explicit depth information. These methods can be classified into three groups: statistical, machine learning-based, and deep-learning-based. In this section, we will provide detailed descriptions of these methods. Table \ref{tbl:point-based-methods} gives a comprehensive overview of point cloud-based methods.

\subsubsection{Statistical}
In this section, we will discuss traditional methods for road defect detection using point clouds. A variety of point cloud processing methods have been used for this purpose. Furthermore, some of the hand-crafted features that have been successfully applied in visual crack detection can also be used in point cloud-based road defect detection.

In recent years, 3D sensors have enabled a plethora of crack detection studies using 3D point cloud data \cite{jiang2018extraction,guan2014iterative,guan2014automated,yu20143d,xu2019intelligent}. For instance, in \cite{jiang2018extraction}, a planar triangulation modelling approach is employed to construct a triangular irregular network dataset and extract cracks based on the inverse distance weighting rasterisation method from wall points. Similarly, \cite{guan2014iterative,guan2014automated} utilise the inverse distance weighting interpolation algorithm, the maximum entropy threshold algorithm, and tensor voting to identify crack curves. To enhance the accuracy of crack analysis and detect cracks from point cloud measurements, \cite{xu2019intelligent} employs the maximum gradient of signal-noise ratio distribution for Gaussian filtering. Bosurgi \etal \cite{bosurgi2023automatic} present a framework for detecting potholes automatically through the
heuristic filtering of depth sensing results of pavement surfaces acquired
using innovative high-performance depth sensors.

Several studies have proposed various methods to detect road defects using 3D point cloud data. In \cite{yu20143d}, the intensity information of point clouds was used to identify pavement cracks and extract their skeletons. 
Li \etal~\cite{li20173d} have presented a clustering-based road defect detection method using 3D point captured by a 3D laser scanner. In~\cite{yang2020intelligent}, a novel method was proposed to identify and extract the cracks, which applied both dilation and the Canny algorithm to the point cloud data of terrestrial laser scanning measurement. In~\cite{tsai2018pothole}, a watershed method was proposed to extract road potholes. Plane fitting is also a widely used technique in point cloud-based road defect detection. For example, in~\cite{du2020pothole}, 3D point clouds were first acquired using binocular stereo vision, and then a pavement plane fitting method was used to eliminate the 3D point clouds of the road plane and roughly extract the pothole. Finally, k-means clustering and region-growing algorithms were adopted to extract the potholes. However, these methods still face challenges in extracting fine and low-connectivity cracks, and the automation of these methods needs to be improved.

\subsubsection{Machine learning-based methods}
Compared with the statistical methods and deep-learning-based methods, conventional machine learning-based methods for trod surface defect detection are relatively less common. This is because the presence or absence of a defect can be determined through the distance information acquired by the laser scanner without needing a separate complex machine-learning algorithm. However, several machine learning-based algorithms have been proposed to develop a more accurate road defect detection model  \cite{wu2016rapid,yan2017detection,yu20143d}.

Wu \etal \cite{wu2016rapid} proposed a novel algorithm for rapidly extracting 3D crack skeletons from 3D point clouds acquired by a mobile LiDAR system. The algorithm leverages a clustering method to distinguish between points belonging to cracks and those belonging to the normal road surface. Similarly, Yan \etal \cite{yan2017detection} use RANSAC to identify points corresponding to road defects. Their method assumes that most scanned points from a road surface can be well-modelled using RANSAC, and the points extracted from road defects can be considered outliers using the RANSAC algorithm. Yu \etal \cite{yu20143d}  also employs a clustering method to gather the 3D point set related to the normal road surface. 

\subsubsection{Deep learning-based methods}
The initial studies on point cloud analysis, such as~\cite{chang2015shapenet, maturana2015voxnet}, typically involved transforming point clouds into regular 3D voxel grids or image grids to enable kernel optimisations and weight sharing. More recently, in~\cite{dizaji20193d}, a 3D InspectionNet was proposed to learn 3D defect features from an artificially generated 3D dataset. This approach can identify defect features from voxels with high accuracy, achieving a validation accuracy of 98.85\%. 

The effectiveness of deep learning in image processing has motivated researchers to apply it to 3D data processing. CrackNet \cite{zhang2017automated}, a deep learning-based method, was proposed to detect cracks in 3D asphalt surface data without using pooling layers to downsample the original data. This approach outputs predicted class scores for individual pixels and has demonstrated high pixel-level accuracy. However, its learning capability is limited by the fixed and non-learnable feature generator. Moreover, the processing speed of CrackNet is slow due to many parameters and significant data depth at hidden layers. Several works have proposed improvements to CrackNet's architecture to address these limitations. CrackNet II \cite{zhang2018deep} abandons the feature generator and constructs a deeper architecture, while CrackNet-V \cite{fei2019pixel} uses small filters to reduce the number of parameters and achieve efficient feature extraction. Additionally, CrackNet-V proposes a new activation unit for detecting shallow cracks. Both CrackNet II and CrackNet-V demonstrate better efficiency and accuracy than CrackNet. However, their improvements are limited and heavily dependent on annotated data. As CrackNet-like approaches are pixel-level techniques, they are still unable to fully exploit 3D information to determine the position of identified cracks accurately.

Conversely, representing the 3D data using the centre of voxels can lead to high memory usage and loss of orientation cues. To address this, some research has used point clouds directly as input, respecting the permutation invariance of points. Feng \etal \etal~\cite{feng2021gcn} applied the Graph Convolutional Network (GCN) for classifying the points. They proposed a graph-widen module to construct a reasonable graph structure for point clouds on GCN to increase the detection performance of road defects from a different perspective.  Li \etal~\cite{li20223d} and Gui \etal \cite{gui2023transfer} modified CNN to process point clouds. Li \etal built a pipeline for accurate segmentation and extraction of road surface objects in 3D LiDAR point clouds. Gui \etal created a framework for cross-scene 3D pavement-crack detection and attribute extraction. Their method was inspired by the heterogeneity of rapid-increasing 3D pavement data and the generalisation ability of transfer learning. PointNet~\cite{qi2017pointnet} was the first to work on the raw point clouds directly, and has been successfully applied in automated defect detection in~\cite{nasrollahi2019concrete} and automated bridge component recognition~\cite{kim2020automated}. 

Although effective for some tasks, PointNet is limited in capturing local structures and recognising fine-grained patterns due to its inability to account for the metric space the points inhabit. PointNet++ was introduced as an extension of PointNet~\cite{qi2017pointnet++} to overcome this limitation. This approach uses a nested partitioning of the input point cloud to learn deep point set features efficiently and robustly. By recursively applying PointNet on these partitions and adaptively combining features from multiple scales, PointNet++ can capture local structures in the point cloud, resulting in improved performance compared to PointNet. For instance, in a pavement disease classification task, PointNet++ outperformed PointNet~\cite{li2021classification}. Chen \etal \cite{chen2023rapid} proposed a deep learning-based gradation extraction method using 3D point cloud data. A multi-feature fusion network was developed, which uses an extraction network as the backbone with additional auxiliary information.

 Point cloud-based methods for road defect detection offer several distinct advantages. These methods provide high-precision 3D data, enabling detailed and accurate mapping of road surfaces and defects. The depth information captured in point clouds allows for a more comprehensive analysis of road conditions, going beyond what is visible on the surface. This depth aspect is particularly useful in identifying subtle or emerging issues that might not be apparent in 2D images. Additionally, using advanced sensors and algorithms, including machine learning and deep learning techniques, enhances the detection and classification of various road defects. These methods can automate the detection process, increasing efficiency and reducing the time and labour involved in manual inspections. Moreover, point cloud-based methods are adaptable to various environments and can work effectively under different lighting and weather conditions, overcoming limitations faced by traditional image-based methods in poor visibility.

However, challenges accompany these benefits. Acquiring and maintaining specialized equipment, such as 3D laser scanners, incurs a high cost. Additionally, processing large volumes of 3D data demands significant computational resources and storage. The resolution of point clouds may pose limitations in detecting very fine or low-connectivity cracks. Automation improvement is needed, especially in handling complex or ambiguous scenarios, where algorithms may struggle, potentially leading to identification or classification errors necessitating manual review or oversight.

Numerous point cloud-based road defect detection methods have been proposed in recent years, utilising both hand-crafted features and deep learning approaches, similar to visual defect perception. However, the current point cloud networks cannot be effectively applied to large scenes or large point clouds due to the high computational load required to extract the latent representation of the point cloud. 
Moreover, the scarcity of public road defect datasets, particularly point cloud data, hinders the development of research on point cloud-based road defect detection. Therefore, the collection of comprehensive 3D road defect benchmark datasets would be beneficial for future studies on road assessment.

\begin{table*}
\caption{List of data fusion-based methods.}
\centering
\resizebox{\textwidth}{!} 
{ 
	\begin{tabular}{c|c|c|c|c|c|m{8cm}}
		\toprule
		Cat & Subcat & Reference & Year  & Based method  & Output protocol  & \multicolumn{1}{c}{Details} \\
		\hline
		
		\multirow{6}{*}[-10ex]{\rotatebox[origin=c]{90}{Machine learning}} & \multirow{3}{*}[-3ex]{\rotatebox[origin=c]{90}{Supervised}}  & Salari \etal \cite{salari2012pavement} & 2012 & SVM  & \makecell{Patch-level\\(Binary)}  &  
		RGB pixel values and the depth information estimated by a stereo-matching algorithm are used as inputs for SVM. 
		\\ \cline{3-7}

         &  & Tsai \etal \cite{tsai2012critical} & 2012 & Scoring method  & \makecell{Pixel-level\\(Binary)}  &  
		A linear-buffered Hausdorff scoring method was used to quantitatively evaluate the crack segmentation performance by comparing the detected cracks with the manually established ground truth. 
		\\ \cline{3-7}
  
        &  & Moazzam \etal \cite{moazzam2013metrology} & 2013 & Probabilistic modelling  & \makecell{Pixel-level\\(Binary)}  &  
		Detect road potholes by analyzing road depth distribution estimated by using an estimated depth map calculated by a stereo matching algorithm.
		\\ \cline{2-7}

    & \multirow{3}{*}[-2ex]{\rotatebox[origin=c]{90}{Unsupervised}}  & Fan \etal \cite{fan2021rethinking} & 2021 & linear iterative clustering   & \makecell{Pixel-level\\(Binary)}  &  
		Estimate a depth map using a stereo vision camera and apply the depth maps and RGB images to derive a model for a normal road surface.
		\\  \cline{3-7}

   &  & Chen \etal \cite{chen2021automatic} & 2021 & Otsu’s algorithm   & \makecell{Pixel-level\\(Binary)} &  
		Proposing an automatic crack-detection method that fuses 3D point clouds and 2D images based on an improved Otsu algorithm.
		\\ \cline{3-7}

   &  & Kim \etal \cite{kim2021crack} & 2021 & RANSAC   & \makecell{Pixel-level\\(Binary)} &  
		A crack identification strategy is proposed, utilising a combination of RGB-D and high-resolution digital cameras to accurately measure cracks irrespective of the viewing angle. The camera system incorporates a custom sensor fusion algorithm dedicated to crack identification, enabling precise measurement resolution and robust depth estimation while addressing the challenge of skewed angles.
		\\ \cline{1-7}

		\multirow{4}{*}[-10ex]{\rotatebox[origin=c]{90}{Deep learning}} & \multirow{4}{*}[-10ex]{\rotatebox[origin=c]{90}{Supervised}}  & Wang \etal \cite{wang2019densefusion} & 2019 & CNN+PointNet   & \makecell{Pixel-level\\(Binary)}  &  
		Image data and point cloud are processed separately, and the processing results are applied to a novel dense fusion network to extract pixel-wise dense feature embedding. 
		\\ \cline{3-7}

   &  & Huang \etal \cite{huang2014pavement} & 2014 & CNN+DNN   & \makecell{Pixel-level\\(Binary)} &  
		A method based on Dempster-Shafer (D-S) theory was proposed to combine the 2D grey-scale image and 3D laser scanning data as a mass function and generate fused detection results for each input image.  
		\\ \cline{3-7}

   &  & Wu \etal \cite{wu2019road} & 2014 & CNN  & \makecell{Pixel-level\\(Binary)} & A new method is proposed for detecting road potholes using mobile point cloud and images. The algorithm has three steps: extracting potential potholes from images using deep learning, extracting candidate potholes from the point cloud, and determining potholes based on depth analysis. Images of potholes and patch distress are used to train and test the deep learning system, as their texture features differ from those of a normal road.
		\\ \cline{3-7}

   &   & Yang \etal \cite{yang2020concrete} & 2020 & CNN+DNN  & \makecell{Patch-level\\(Binary)}  &  
		RGB images were first used to in-paint per depth frame, and then a depth adaptive sliding-window size selection method was used to adjust the bounding box size. 
		\\ \cline{3-7}

       &  & Bolourian \etal \cite{bolourian2023point} & 2023 & CNN,PointNet++  & \makecell{Point-level\\(Binary-class)} &  
		developed a point cloud–based semantic segmentation DL method (SNEPointNet++) to detect different types of concrete surface defects. Use data fusion between the point cloud and images transferred from the point cloud data.
		\\ \cline{1-7}

		\bottomrule
	\end{tabular}
	}
	\label{tbl:fusion-based-methods}
%\vspace{-1em}
\end{table*}

\subsection{RGB+D data fusion-based methods}
As discussed in the above sections, image-based methods rely on visual cues such as texture, colour, or brightness to identify road defects. However, they may struggle to detect defects that are not clearly visible in the image, such as those covered by shadows or occluded by other objects. On the other hand, explicit depth information allows for identifying road defects that are not visible in the image alone. Various methods have been proposed to combine image and depth information. This survey paper categorises these methods into machine learning-based and deep-learning-based methods. This section presents detailed descriptions of these methods using image and depth information, listed in Table \ref{tbl:fusion-based-methods}. 

\subsubsection{Machine learning-based methods}
RGB images and depth maps have their own advantages and limitations when it comes to road defect detection. RGB images can be helpful for capturing information about the texture and shape of defects on the road surface. However, depth maps may not be as effective in conveying precise information about surface texture. Nevertheless, depth maps are valuable in providing general defect detection performance by offering depth information, which can be critical in identifying defects.

Recent studies have proposed various machine learning methods that leverage the strengths and weaknesses of different data sources \cite{salari2012pavement,moazzam2013metrology,fan2021rethinking,tsai2018pothole}. Tsai \etal~\cite{tsai2018pothole} applied a simple filtering technique on the colour histogram to find road defects on asphalt roads. Chen \etal~\cite{chen2021automatic} leveraged a thresholding algorithm to detect road cracks using point clouds and RGB images. Salari \etal \cite{salari2012pavement} applied SVM to classify road defects in small-size image patches. Their approach utilised RGB pixel values and depth information obtained from stereo-matching algorithms as inputs to the SVM. Moazzam \etal \cite{moazzam2013metrology} focused on detecting road potholes by analysing the depth distribution of roads, which was estimated by using stereo matching algorithms. Kim \etal~\cite{kim2021crack}  leveraged the RANSAC algorithm to identify road cracks based on depth information, and they used image information to assist in finding valid locations of the cracks.

Fan \etal \cite{fan2021rethinking} used a stereo vision camera to estimate depth maps and combined them to develop a model for a normal road surface. They applied a simple linear iterative clustering algorithm to derive the model. 

\subsubsection{Deep learning-based methods}
The fusion of vision and depth data has emerged as a prominent topic in the field of computer vision and robotics. This approach has been employed in numerous applications, including 2D/3D object detection~\cite{hoffman2016cross, xu2018pointfusion}, object segmentation~\cite{gupta2014learning}, and object pose estimation~\cite{xiang2017posecnn, wang2019densefusion}. By leveraging both visual and depth information, these applications aim to improve the accuracy and robustness of their outputs, leading to more effective and reliable results.
% compare the difference between vision and depth.

In the past, most studies treated depth information as an image Grid and used the same type of feature learner for visual and depth data. However, this approach fails to fully leverage the potential of depth information, even when the depth information is represented with pseudo colours~\cite{hoffman2016cross} or geometrically codes~\cite{gupta2014learning}. This limited use of depth data can result in suboptimal performance, particularly when dealing with complex scenes or objects with complex geometries.

Recent studies have highlighted the advantages of using heterogeneous network architectures to fuse visual and depth information to avoid the lossy input pre-processing such as projection or quantisation~\cite{xu2018pointfusion, wang2019densefusion}. For example, in~\cite{xu2018pointfusion}, a CNN and PointNet were used independently to process the image data and raw point cloud data, respectively. The resulting image and point cloud representations were then fused to predict 3D boxes. Similarly, in~\cite{wang2019densefusion}, the image data and point cloud were also processed individually, and a novel dense fusion network was introduced to extract pixel-wise dense feature embedding. Furthermore, in identifying road defects, the synergy of visual and depth perception can notably boost detection accuracy. Attributes from visual data, like colour, can expedite recognising potential defect-prone areas. On the other hand, depth data adds resilience to the detection process, is minimally influenced by factors like illumination and disturbances, and adeptly maps the authentic size and shape of defects. By harnessing the combined potential of CNN and PointNet, it is evident that diverse network structures can amplify the consistency and precision of fusion-centric techniques.

Despite the potential benefits of fusing vision and depth data for defect detection, there are only a few methods that have explored this approach in a straightforward manner. For example, Huang \etal \cite{huang2014pavement} proposed a method that uses the Dempster-Shafer theory to combine 2D grayscale images and 3D laser scanning data as a mass function, and generates fused detection results at the decision-making level. Wu \etal \cite{wu2019road} proposed a pothole detection method based on a deep learning-based feature ensemble for the two features extracted from the point cloud and RGB image. 

In a more recent study, Yang \etal \cite{yang2020concrete} used RGB images to paint per depth frame, and then applied a depth adaptive sliding-window size selection method to adjust the bounding box size. However, there is a growing interest in end-to-end heterogeneous network architectures, such as DenseFusion \cite{wang2019densefusion}, which can offer a more effective and efficient means of integrating vision and depth data for defect detection. By leveraging the strengths of these advanced fusion-based methods, researchers can make significant progress towards developing more robust and accurate defect detection systems.

 RGB+D data fusion-based methods for road defect detection leverage the strengths of RGB imagery and depth data, providing a range of advantages. The primary benefit is the comprehensive analysis capability: RGB images provide detailed information about surface texture and colour, aiding in recognising defects like cracks or potholes. Depth data complements this by providing crucial three-dimensional information about the road surface, particularly useful for identifying defects that are not easily discernible in two-dimensional images. This synergy enhances the overall accuracy and robustness of defect detection, making the system less susceptible to issues like poor lighting or visual obstructions. Additionally, these methods can better understand defects' actual size and shape, an aspect critical for effective road maintenance. The fusion approach also adapts to various environmental conditions, offering consistent performance under different lighting and weather scenarios.

However, there are notable drawbacks to RGB+D data fusion methods in road defect detection. One of the primary challenges is the complexity and computational intensity of processing and fusing two distinct types of data, which requires advanced algorithms and significant processing power. This complexity may result in higher hardware and software costs and the requirement for specialised expertise in developing and maintaining these systems.

\subsection{Evaluation metrics for Defect Detection Methods}

In road defect detection, the selection of evaluation metrics depends on the methodologies to identify the road defects that are tailored to the two types of classification tasks: one-class classification (\aka anomaly detection) \cite{yu2018joint,yu2023multi,oliveira2012automatic,abdel2006pca,oliveira2010improved,oliveira2009automatic,buza2013pothole} and multi-class classification \cite{li2021classification,fang2020novel}. The evaluation metrics for one-class classification measure methods' performances to predict whether new data is normal or an anomaly, including unsupervised or semi-supervised approaches, which usually train their models with normal samples only (\ie, one-class). On the other hand, multi-class evaluation metrics are applied to evaluate the performance of multi-class road defect detection \cite{redmon2018yolov3,liu2016ssd} or instance segmentation \cite{wang2020solov2,bolya2019yolact}.

For one-class classification tasks commonly seen in road defect detection based on unsupervised or semi-supervised methods, the focus predominantly distinguishes road defects from normal road surfaces. The metrics commonly used include the False Positive Rate (FPR), which indicates the rate at which normal instances are incorrectly classified as anomalies, and the True Positive Rate (TPR), which measures the accuracy in identifying actual anomalies \cite{yu2021normality,pidhorskyi2018generative}. The Receiver Operating Characteristic (ROC) Curve and the Area Under the Curve (AUC) are also crucial in these scenarios \cite{yu2018joint,yu2023multi,abdel2006pca,oliveira2010improved}, as they provide insights into the performance across various threshold settings. The Precision-Recall Curve gains importance in datasets with a severe imbalance between normal and anomalous instances\cite{oliveira2009automatic,buza2013pothole}.

In the case of multi-class road defect detection and classification, considering instance segmentation, metrics like Precision, Recall (or Sensitivity), and the F1-Score become more significant \cite{li2021classification,fang2020novel}, because the metrics should provide the performance evaluation for detection results on multiple road defect classes. Precision measures the proportion of true positive detections among all positive detections for each defect class, ensuring the accuracy of defect identification. Recall assesses how many defects of a particular type are correctly identified, which is crucial for not missing any critical defects. The F1-Score balances precision and recall, which is especially valuable when the dataset is imbalanced. In addition, the Mean Absolute Error (MAE) and Intersection over Union (IoU) are employed in scenarios where the exact localisation of each defect type is significant \cite{fang2020novel}. Accuracy, while a common metric, can be less informative in multi-class scenarios with imbalanced classes 
\cite{li2021classification}.

These metrics, tailored to the specific nature of the classification task, are fundamental in evaluating and enhancing the performance of road defect detection systems. They provide a comprehensive understanding of how well these systems can distinguish between different classes of defects or identify anomalies in various operational conditions.

%% file: 05_Conclusion.tex
\section{Conclusion}
\label{sec:6}
This survey paper comprehensively explores the current landscape in road defect detection, from traditional image-based methods to the emerging realms of point cloud and data fusion techniques. While image-based methods have seen significant advancements and remain a primary focus, our review highlights a burgeoning interest in point cloud-based methods and data fusion approaches. These emerging methodologies hold immense potential for revolutionising road defect detection, although they still need more exploration. 

The point cloud and data fusion-based methods open a new avenue for future research. Point cloud data has different geometrical properties than image domains, and it may not be suitable to process the data using the existing image-based methods. Data fusion between the two different domains is also quite a complicated issue. Hence, developing more sophisticated algorithms capable of processing vast and complex datasets will be crucial in advancing these technologies.

Applying these advanced methods in road defect detection can significantly contribute to smarter public transportation infrastructure management. With the integration of IoT and smart city initiatives, these technologies can provide real-time, accurate road condition data, facilitating proactive maintenance and enhancing public safety. Moreover, the environmental impact of road maintenance can be optimised by employing these precise detection methods, aligning with sustainability goals.

In conclusion, while the field has made remarkable strides in image-based road defect detection, the untapped potential of point cloud and data fusion methods presents exciting prospects. These technologies are not just evolutionary steps in road assessment but could be pivotal in shaping future intelligent transportation systems and sustainable infrastructure management. The road ahead, paved with these advanced technologies, beckons further exploration and innovation.